\documentclass[aps,pre,reprint,twocolumn,superscriptaddress,nofootinbib]{revtex4-2}

\usepackage[utf8]{inputenc}
\usepackage[T1]{fontenc}
\usepackage{hyperref}
\hypersetup{hidelinks}
\usepackage{url}
\usepackage{graphicx}
\usepackage{amsmath,amssymb,amsfonts}
\usepackage{nicefrac}
\usepackage{xcolor}
\usepackage{microtype}
\usepackage{array}
\usepackage{booktabs}
\usepackage{longtable}

\definecolor{crimson}{RGB}{165,28,48}
\hypersetup{colorlinks=true,allcolors=crimson}
\usepackage{XCharter}

\newcommand{\ttcell}[2]{%
  \parbox[t]{#1}{\ttfamily\raggedright #2}%
}

\newcommand{\correspondencenote}{%
  \thanks{Contact: \mbox{fpasha@fas.harvard.edu};
  \mbox{vnmurthy@fas.harvard.edu}}%
}

\begin{document}
\title{Learning Perceptual Geometry of Natural Odor Mixtures \texorpdfstring{ \\}{} from LLM-Derived Descriptors}

\author{Farhad Pashakhanloo}
\correspondencenote
\affiliation{Center for Brain Science, Harvard University}

\author{Venkatesh N. Murthy}
\correspondencenote
\affiliation{Center for Brain Science, Harvard University}
\affiliation{Department of Molecular and Cellular Biology, Harvard University}
\affiliation{Kempner Institute for the Study of Natural and Artificial Intelligence, Harvard University}

\begin{abstract}
The relationship between odor chemistry and perception remains incompletely understood. Many existing studies focus on the perception of single molecules, whereas natural odors contain tens to hundreds of components, adding extra layer of complexity to percept formation. Here, we explore the relationship between molecular composition and perceived similarity in natural odor objects. We study this using a database of more than 1,200 volatile-compound-in-food mixtures containing more than 8,000 molecular components, and use large language models (LLMs) as a scalable surrogate for deriving mixture-level perceptual descriptions. We find that language-derived representations provide useful supervision
for predicting held-out mixture similarities from molecular composition. The prediction was consistently strongest when perceptual targets were derived from chemosensory prompts than from other sensory prompts. 
Further, during the mixture-level training, we learned low-rank molecular embeddings that enabled better prediction of held-out mixture similarities
than component-overlap metrics, even for mixture pairs with no shared molecules.
Although no single-molecule labels were used during training, the learned molecular geometry showed significant alignment with the independent human odor-descriptor space. Finally, we explored the use of external molecular features in our
mixture-similarity prediction task. These features showed modest zero-shot alignment with mixture-level
perceptual similarity. They improved prediction under limited supervision
but offered little additional predictive benefit with the full training set. Together, these results show that language-derived descriptions can
provide scalable supervision for predicting the perceptual organization of complex natural mixtures, contributing to a quantitative understanding
of how molecular composition gives rise to odor perception.
\end{abstract}

\maketitle 

\section{Introduction}
Natural odors are complex mixtures typically containing tens to hundreds of volatile molecules. The perception of a natural mixture depends both on the odor qualities associated with its individual molecular components and on how those components combine to produce the overall percept \cite{doty2025odors, weiss2012olfactorywhite, dravnieks1985atlas, ohloff2012scent, schiffman1974physicochemical, howard2014configural, thomasdanguin2014perception, berglund1976psychological,stevenson2007odour, bushdid2014humans}.

Even at the level of individual molecules, the mapping from molecular structure to perceived odor remains incompletely understood. Simple measures of chemical similarity do not fully capture similarities in perceived odor \citep{keller2016diverse, keller2017predicting, gutierrez2018natural}. Machine-learning approaches, including convolutional neural networks, graph neural networks, and transformer-based molecular representations, have been used to predict odor descriptors and perceptual similarities for individual molecules \citep{shuvaev2024deepnose,lee2023pom,ross2022molformer,taleb2024transformers}.

Mixtures introduce a further level of complexity because it has long been unclear whether their perceptual qualities can be predicted by linearly combining those of their constituent molecules or require a more complex mapping from components to mixture perception \citep{snitz2013similarity}. More generally, there is an unresolved tension between analytical and synthetic processing of odor mixtures \citep{jinks2001analysis,rokni2014olfactory, howard2014configural, thomasdanguin2014perception, berglund1976psychological, stevenson2007odour}. Recent evidence suggests that the perceptual profiles of controlled mixtures containing two to ten components can largely be explained by linear averaging of their component profiles \citep{pellegrino2026odors}. Related work has developed mixture-similarity models using inductive biases over molecular sets and human pairwise ratings \citep{tom2025mixtures}. Whether these aggregation principles extend to natural odors containing tens to hundreds of reported components remains unclear, especially since nonlinear interactions are ubiquitous even at the first stage of neural encoding \cite{zak2020antagonistic, singh2019competitive, pfister2020odorant, inagaki2020widespread}.

One difficulty in relating molecular composition to odor is quantifying perception itself. Language provides a natural means of grounding perceptual experience in descriptions that can be communicated and compared \citep{iatropoulos2018language,horberg2022semantic, majid2018olfactory}.  Accordingly, psychophysical studies use language by asking participants to characterize odors using fixed vocabularies of perceptual descriptors \citep{dravnieks1985atlas}. Complementary approaches quantify perceptual organization more directly through pairwise odor-similarity judgments \citep{snitz2013similarity}. Such measurements have enabled quantitative analyses of olfactory perceptual space, revealing low-dimensional, hierarchical, and non-Euclidean organization \citep{koulakov2011olfactoryspace,zhou2018hyperbolic,yang2023fly,sajan2026hierarchies,medbouhi2026geometric}. 
However, grounding odor perception in language presents an important challenge. Odors lack a universal descriptive vocabulary, so similar experiences may be expressed using different words, while fixed descriptor lists restrict responses to qualities chosen in advance.

Large language models offer a scalable way to extend this language-based approach beyond fixed descriptor vocabularies.
Their outputs have been shown to capture aspects of human sensory judgments across several modalities, and recent benchmarking indicates that they contain substantial knowledge about single-molecule odorants and their perceptual qualities \citep{marjieh2024llm,makri2026benchmark}.
The field of olfaction could particularly benefit from using LLMs to describe perceptual qualities.
This is because psychophysical measurements of complex odor mixtures are scarce, and the richness of olfactory perceptual space makes comprehensive human characterization difficult at scale. Natural odor sources are particularly suitable for this approach because references to them are likely to occur across sensory, culinary, ecological, and everyday contexts in language-model pretraining corpora, giving the models richer contextual associations with their sensory qualities than are typically available for isolated volatile compounds.

Here, we investigate how the molecular composition of complex natural
odors relates to their perceptual similarity. We augment a database of
natural-mixture compositions with surrogate odor descriptions generated
by a large language model. We then use the resulting representations
to learn a composition-based mixture-similarity metric parameterized by
low-rank molecular embeddings. We evaluate its predictive performance
across sensory and semantic conditions and compare the learned molecular
geometry with independent human odor annotations. We further test
whether external single-molecule features improve prediction and how
their contribution depends on the amount of mixture-level supervision.
Our results show that language-derived supervision can support prediction
of the perceptual organization of complex natural mixtures while
recovering both odor-relevant and broader sensory and semantic structure.

\section{Results}

\subsection{Using language models to characterize the perceptual space of natural mixtures} \label{ResultsA}

\begin{figure*}[t]
    \centering
\includegraphics[width=1\linewidth]{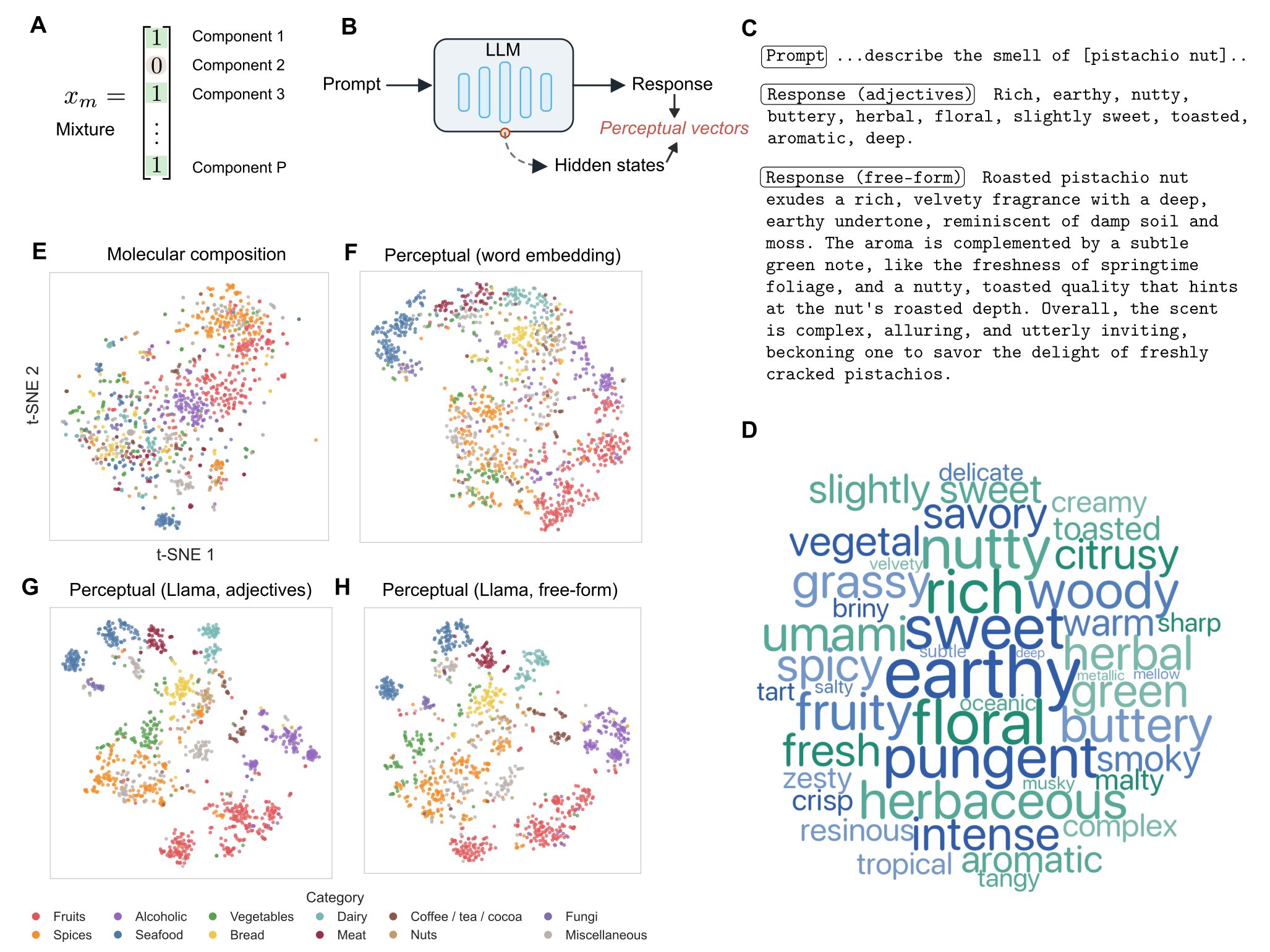}
    \caption{\textbf{Natural-mixture dataset and language-derived representations.} (A) Each mixture can be represented by a binary vector $x_m \in \{0,1\}^P$ specifying the presence or absence of different molecules. (B) Llama-3.1-8B-Instruct was prompted to describe each mixture using fixed adjective lists or free-form text, and corresponding perceptual targets were created from hidden state activations or word-embeddings. (C) Example prompt-responses of the model (see Table \ref{tab:prompt-output-examples} for complete prompts). (D) Smell descriptor word cloud. The 45 most frequent descriptors pooled across five generation runs from the 10 adjectives pass; word size indicates frequency. (E--H) Two-dimensional t-SNE visualizations of mixture representations
based on (E) binary molecular composition vectors, (F) pretrained word
embeddings of 10 generated odor adjectives, (G) Llama activations from
10-adjective responses, and (H) Llama activations from free-form odor
descriptions. All t-SNE projections were computed using pairwise cosine
distances. Each point represents a mixture and is colored by food category. An interactive version of panels~(F--H) is available
\href{https://vnmurthylab.github.io/smellm/widget/}{online}.}
    \label{fig1}
\end{figure*}

We used a dataset of approximately 1,200 natural mixtures, compiled from Volatile Compounds in Food (VCF) Online \citep{vcfonline2026}. Each entry corresponds to a common food source, such as a fruit, spice, or beverage, and lists the volatile molecular components reported for that source. This entry is paired with a $P-$dimensional binary vector that specifies the volatile molecular compounds constituents of that entry: a value of 1 indicates that the component was reported for that source, and 0 indicates that it was not listed ($P \approx 8500$, Fig.~\ref{fig1}A). To obtain a smell description for each mixture, we prompt an LLM (\texttt{Llama-3.1-8B-Instruct}) with the name of each entry, and ask it to return either a specified number of odor adjectives or a free-form paragraph describing the smell of that mixture (Fig.~\ref{fig1}B,C). In both cases, the model generates descriptions without being restricted to a predefined vocabulary. In the case of adjectives response, we ask for a predefined number of adjectives to be included in the response ($10$ by default but varied in parameter sweeps). Across all the mixtures, a single generation pass of the $10$-adjective condition yielded more than \(400\) unique smell descriptors. Fig.~\ref{fig1}D shows word clouds of the most frequently generated
adjectives. The descriptors spanned diverse odor qualities and included
terms also associated with taste, highlighting overlap in the language
used to describe these senses.
We convert these adjective lists and free-form responses into three perceptual target representations. For both cases, we pool the generated-token hidden-state activations from layer 18, yielding $4096$-dimensional Llama-adjective and Llama-free-form targets. For the adjective responses, we additionally map each generated adjective to its pretrained GloVe embedding (\texttt{glove-wiki-gigaword-300}) and average the resulting vectors, yielding a $300$-dimensional adjective word-embedding target. Thus, our three target constructions comprise  word-embeddings of adjective lists, Llama activations from adjective lists, and Llama activations from free-form descriptions. Fig.~\ref{fig1}(E--H) show two-dimensional t-distributed stochastic neighbor embedding (t-SNE) projections of mixture representations in the composition space (panel E), as well as the three LLM-derived perceptual embeddings (panels F--H). These projections suggest qualitatively different patterns of local clustering across the molecular-composition and language-derived representation spaces.

\subsection{Predicting perceptual similarity from mixture components.}

The use of language models allowed us to augment the molecular composition of each mixture with a language-derived surrogate of its perceived odor. This led to a paired composition-perception dataset in which every mixture is represented by both a binary molecular components vector (binary, indicating only the presence of components) and a perceptual target vector. We next examined how well perceptual similarity in this space could be predicted from the mixture composition.

\subsubsection{Alignment between composition and perceptual space}
Let $X\in\{0,1\}^{N\times P}$ be the binary mixture-by-molecule matrix, where $N$ is the number of mixtures and $P$ is the number of distinct molecular components. We denote by $x_m\in\{0,1\}^P$ the binary component vector for mixture $m$. For each of the three target constructions described above, the language-model pipeline produces a perceptual target vector $y_m\in\mathbb{R}^d$, where $d=4096$ for the Llama hidden-state targets and $d=300$ for the word-embedding adjective target. 
We frame the prediction problem here at the level of similarity among mixtures, and ask whether mixture composition can recover a relational organization of the perceptual space. An alternative but closely related formulation aims at predicting each perceptual target vector directly, $\hat{y}_m=f(x_m)$; we discuss this approach in Appendix~\ref{sub:ridgeregression} and return to it later when evaluating external molecular features. Characterizing the relational organization requires specifying how pairs of mixtures are compared in each space. For the language-derived perceptual representations, we define the similarity between mixtures $m$ and $n$ as
\[ S^{\mathrm{perc}}_{mn} = k^{\mathrm{perc}} (y_m, y_n)  = 
\cos(y_m,y_n),\]
where $k^{\mathrm{perc}}$ denotes the similarity kernel in the perceptual target space. We use cosine similarity because it is a standard measure of angular proximity in word-embedding and language-model representation spaces \citep{pennington2014glove,park2023linear}.
Likewise, a similarity matrix can be defined in the mixture composition space,
\[ S^{\mathrm{comp}}_{mn} = k^{\mathrm{comp}} (x_m, x_n).\]
Alignment between the mixture-composition and perceptual spaces can then be quantified using representational similarity analysis (RSA) \citep{kriegeskorte2008rsa} by computing the Pearson correlation between the corresponding upper-triangular entries of the representational
similarity matrices (RSMs) $S^{\mathrm{comp}}$ and $S^{\mathrm{perc}}$.

\subsubsection{Choice of Metric in composition space}
A key question in understanding the mapping from components to perception is how similarity should be defined in the composition space. 
Different choices of $k^{\mathrm{comp}}$ reflect different assumptions about which aspects of molecular composition are perceptually relevant.
A simple choice would be to define it as a measure of direct component overlaps between two mixtures. Two such choices are the cosine similarity between the binary vectors of the mixtures
or the Jaccard metric which is a normalized measure of the component overlap. 
An issue with both of these metrics is that they vanish if two mixtures have no shared components, even though they may contain distinct molecules that have similar perceptual tone. Additionally, when comparing two mixtures, both metrics weight all the molecules equally, which is known to be unrealistic.

\begin{figure*}[t]
    \centering
    \includegraphics[width=0.7\linewidth]{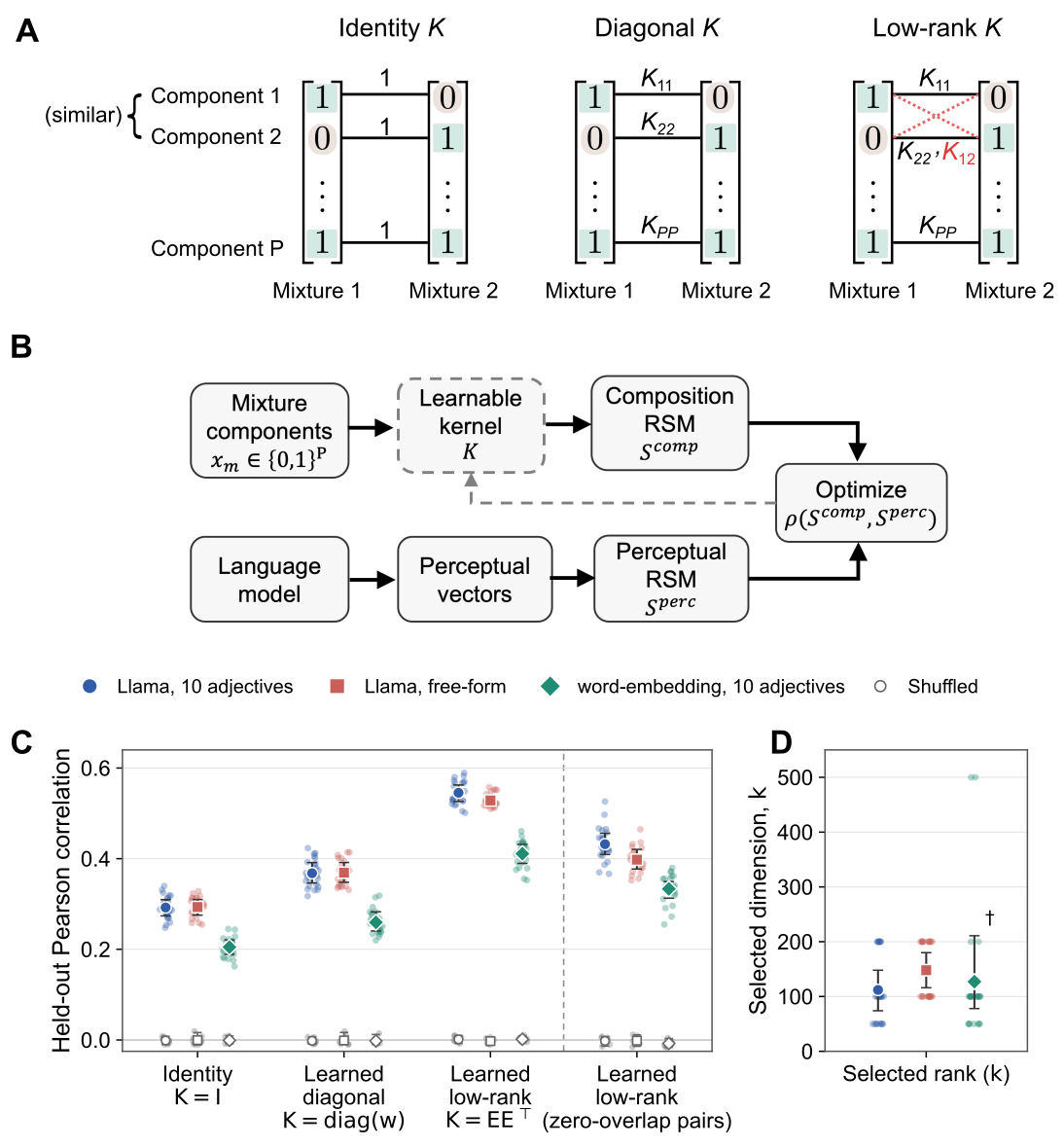}
    \caption{\textbf{Predicting perceptual similarity from mixture compositions.} (A) Schematics showing different kernels for comparing two mixtures. We show an example in which components 1 and 2 have similar embeddings, however, the mixture do not share those components. The identity and the diagonal kernels do not capture this effect leading to a lower similarity between mixtures. However, the kernel on the right can capture this effect due to the non-zero off-diagonal term  $x_{a1}x_{b2}K_{12}$. Note that high similarity of the components lead to a high value of $K_{12} = e_1^{\top}e_2$. (B) Similarity-matching model. The perceptual branch prompts an LLM with mixture context, embeds generated descriptions, and forms a target perceptual RSM. The molecular branch maps mixture composition to a model RSM; the dashed module is a learned molecule embedding. (C) Held-out RSA correlation, $\rho\left(S^{\mathrm{comp}}_{\mathrm{test}},S^{\mathrm{perc}}_{\mathrm{test}}\right)$, for three types of kernels. The colors represent the source of perceptual targets. Shuffled values are calculated from randomly permuting the perceptual data before training and then calculating the held out RSA correlations. In the case of diagonal and low-rank kernels, this means the kernels were retrained with shuffled labels. (D) Validation-selected dimension $k$ for the low-rank kernel fits. $\dagger\!$ One fit with $k=1000$ was excluded.
    Bars show means across runs and splits and error bars indicate 95$\%$ CI. \vspace{0em}}
    \label{fig2}
\end{figure*}

To move beyond direct component overlap, here we use a more flexible framework that leads to a Mahalanobis-type metric in the mixture composition space. Consider an
embedding matrix
$E\in\mathbb{R}^{P\times k}$, where each row contains a $k-$dimensional embedding vector for a given component. 
A mixture with binary component vector $x_m$, can be represented in this embedding space by linear aggregation of its component
\begin{equation*}
    z_m=\frac{1}{n_m}E^{\top}x_m
    \in\mathbb{R}^k,
\end{equation*}
where $n_m$ is the number of molecular components in mixture $m$.
Now, if we define the mixture similarity as the cosine similarity between their mixture representations, the resulting composition kernel is
\begin{align} \label{cossimzmzn}
k^{\mathrm{comp}}_K(x_m,x_n)  
&= \cos(z_m,z_n) \nonumber \\
&=
\frac{x_m^\top Kx_n}
{\sqrt{
\left(x_m^\top Kx_m\right)
\left(x_n^\top Kx_n\right)
}},
\end{align}
where
\[ K = E E^{\top} \succeq 0\]
is a positive semidefinite component-by-component kernel matrix satisfying \(\operatorname{rank}(K) \leq k < P\). To get an intuition on how elements of kernel $K$ contribute to the similarity function between two mixtures, the numerator of the fraction in Eq.~\ref{cossimzmzn} can be expanded as:
\begin{equation} \label{sumK}
x_m^{\top}K x_n
=
\sum_{a=1}^{P}\sum_{b=1}^{P}
x_{ma}x_{nb}K_{ab}.
\end{equation}
This suggests the similarity between two mixtures depends not only on shared components (\(a=b\)), but also on pairs of distinct components (\(a\neq b\)), weighted by \(K_{ab}=e_a^{\top}e_b\), where \(e_a^{\top}\) denotes the \(a\) row of the embedding matrix \(E\). Below, we consider three increasingly more general cases of $K$ (see also Fig.~\ref{fig2}).
\paragraph{Identity kernel ($K=I$):} Here, the summation will be limited to overlapping components with equal weights, and the kernel becomes identical to the binary cosine kernel measure, i.e. $K^{\mathrm{comp}}(x_m,x_n) = \cos(x_m,x_n).$
\paragraph{Diagonal kernel ($K=\operatorname{diag}(\{K_{aa}\})$):} This case can be represented by an embedding matrix $E \in \mathbb R^{P \times P}$ with mutually orthogonal rows. Here, too, only the overlapping indices contribute to the summation, but with different weights assigned to each component.
\paragraph{Low-rank kernel:} 
This case allows cross-molecular relationships through \(K = EE^\top\), where \(E \in \mathbb{R}^{P \times k}\) and \(k < P\). Notably, if components $1$ and $2$ have similar perceptions, their embeddings tend to align, increasing the off-diagonal term $K_{12} = e_1^{\top}e_2>0$. This appears in the sum of Eq.~\ref{sumK} as $(x_{m1}x_{n2} + x_{m2}x_{n1})e_1^{\top}e_2$, suggesting that the presence of either component is sufficient to increase the similarity between the mixtures. Fig.~\ref{fig2}(A) demonstrates how this model can capture relationships between distinct molecular components. 

\subsubsection{Learning the similarity metric from data}
The low-rank formulation above defines a flexible family of similarity functions that can account for relationships among distinct molecular components. Here, we test whether such a function can be learned from the training data to reproduce the perceptual organization of mixtures. Specifically, we optimize the embedding matrix $E$ to align the composition similarity matrix $S^{\mathrm{comp}}$ with its perceptual counterpart $S^{\mathrm{perc}}$ (Fig.~\ref{fig2}B):
\begin{equation}
\hat{E}
=
\underset{E}{\operatorname{arg\,min}}
\left[
-\rho\!\left(S^{\mathrm{comp}}(E),
S^{\mathrm{perc}}\right)
+\lambda R(E)\right].
\label{eq:optimization}
\end{equation}
In the above, $\rho$ denotes the Pearson correlation between the upper-triangular entries of the two matrices, and $R(E)$ is a regularization term.
We partition the mixtures into 60\% train, 20\% validation, and 20\% test sets randomly, and compute the corresponding similarity matrices for each group independently.
We use the validation data to choose the embedding dimension $k$, early stopping and regularization parameters, and report the held-out (test) correlation as a measure of alignment. Optimization is performed using a full-batch gradient descent with Adam on the training data.

Fig.~\ref{fig2}(C) summarizes the held-out correlations across different kernels as well as different perceptual targets. We see that diagonal kernel performance is better than that of a fixed identity kernel, independent of the type of perceptual target (for Llama, 10 adjectives, identity: \(r=0.292\,[0.274,\,0.310]\); learned diagonal: \(r=0.368\,[0.346,\,0.391]\); learned low-rank: \(r=0.545\,[0.526,\,0.563]\); values in brackets indicate crossed-bootstrap 95\% CIs). This suggests weighting the molecules differently, as is done in the learned diagonal kernel, is not enough for accounting for the perceptual similarities and cross-molecular terms play a substantial role. 
Intuitively, the low-rank model can assign similar embeddings to different molecules that make similar contributions to the language-derived perceptual organization, allowing mixtures to be similar even when they do not contain the same components.

To inspect this directly, and to isolate what the low-rank kernel adds beyond molecular overlap, we considered test pairs for which there is no molecular overlap. For these pairs, the identity and the diagonal kernels assign zero similarity by design and therefore cannot distinguish among them. In contrast, the learned low-rank kernel remained correlated with perceptual targets, with $r=0.432\pm0.012$ across approximately $9{,}800$ zero-overlap pairs, compared with $r=0.545\pm0.009$ across approximately $30{,}600$ test pairs overall for the Llama 10-adjective target. Thus, the learned cross-molecular relationships generalized to held-out combinations even in the absence of direct component overlap. These results indicate that, when supported by sufficient evidence in the training data, the model can infer that distinct molecules play similar roles in organizing the language-derived perceptual space.

During training, we selected the kernel dimension $k$ by sweeping candidate values and maximizing performance on the validation data. Across runs, the selected dimension was consistently at least $50$, with a pooled mean of $k=141$ (95\% CI: $[99,200]$) across all fits and target representations (Fig.~\ref{fig2}D). As $k$ is the dimensionality of the learned molecular embedding, it defines a bottleneck between the $P$-dimensional component space and the learned similarity space. Notably, the selected dimensionality is of the same order of magnitude as the number of human odorant-receptor types, which form a biological bottleneck that compresses the vast chemical space of volatile molecules into signals underlying odor perception.

\subsection{Alignment between learned molecular geometry and independent odor annotations}
\begin{figure*}[t]
    \centering
    \includegraphics[width=1\linewidth]{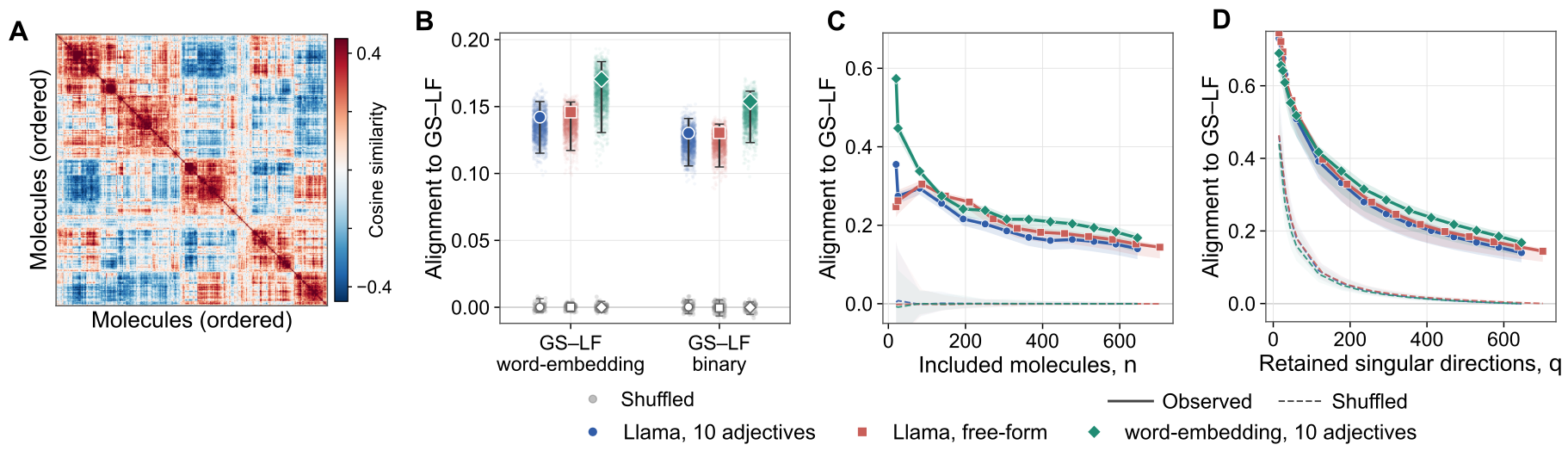}
    \caption{\textbf{Learned molecular geometry and alignment with external molecular features.}
    (A) Example learned molecule--molecule cosine similarity matrix, using a subset of the most frequent training-set molecules ordered by hierarchical clustering 
    (B) Alignment to GS-LF data across different prompts. In calculating the alignment scores, perceptual RSM for the GS-LF data are derived either from word embedding averaging or from binary overlap of the available descriptors between two molecules. Gray points denote alignments after permuting the correspondence between molecules in the learned and GS--LF representations. A separate control that shuffled perceptual labels before model training produced a similarly near-zero alignment.
    (C) Alignment to GS-LF after filtering molecules based on their geometric stability across fits. The threshold on stability is varied leading to a lower number of included molecules from right to left. (D) Alignment to GS-LF after projection of RSMs to restricted subspaces. Moving from right to left progressively restricts the representations to better-conditioned directions.
    In (C) and (D),
    shading denotes crossed-bootstrap $95 \%$ CI for the observed curves and $95 \%$ permutation-null intervals around the dashed shuffled curves. In all the panels, we included molecules with prevalence of $\geq 10$ across mixtures, and a consensus RSM was created from all the fits before alignment.}
    \label{fig:alignment}
\end{figure*}
So far, we have asked whether molecular composition could recover language-derived similarities among natural mixtures. These results evaluate the model at the level of whole mixtures. In this process, a low-rank kernel is learned which induces a molecule-to-molecule similarity geometry, despite receiving no direct single-molecule supervision. An example of this similarity kernel is shown in Fig.~\ref{fig:alignment}(A).
A natural next question is whether this learned single-molecule organization is meaningful. In other words, do molecules placed near one another share known molecular properties? We first addressed this question by comparing the learned molecular geometry with that derived from human odor annotations from the independent GoodScents–Leffingwell (GS–LF) database, and curated by \citep{lee2023pom}. To do this, we first found common molecules between the two databases and constructed RSMs over the matching molecules in both spaces. The GS–LF dataset represents each molecule by a fixed vocabulary of 137 odor descriptors, with each descriptor recorded as either present or absent for a given molecule. One way to define the similarity of two molecules is to directly compare these binary annotations, which would measure how many descriptor words two molecules share. However, it would treat related terms such as “citrus” and “lemon” as entirely distinct. To account for these semantic relationships, we mapped each assigned descriptor to a pretrained word embedding and averaged the embeddings associated with each molecule. We then defined the similarity between two molecules as the cosine similarity between their resulting representations.

Although the model was trained using mixture-level supervision, the resulting single-molecule geometries showed measurable alignment with the independently constructed GS--LF odor-descriptor space. This is shown in Fig.~\ref{fig:alignment}(B) across perceptual targets, and with the binary and word-embedding similarities on the GS-LF adjectives as described above.

The result above considers the overall alignment between the geometries of hundreds of molecules between the two spaces. However, the representation of individual molecules are not all estimated with equal reliability. This is because the model encounters molecules only through the mixture--composition design matrix. As a result, the relationships involving sparse or redundant occurrence patterns are less strongly constrained and may vary across fits (see
Appendix~\ref{sec:identifiability} for a formal analysis). We therefore repeated the GS--LF comparison while progressively focusing on portions of the molecular geometry that had a more stable estimation. Specifically, we first calculated the empirical stability of individual molecules across fits from multiple runs, and then measured the alignment limited to the molecules with a minimum stability. Retaining progressively more stable molecules produced a marked increase in GS--LF alignment to \(r=0.57\) for the 20 most stable molecules, while the molecule-shuffled control remained near zero (Fig.~\ref{fig:alignment}C).
We also performed a similar analysis by restricting the alignment to progressively higher subspace of the mixture--composition matrix. Restricting the comparison to the best-observed directions of the composition matrix increased the alignment to approximately \(r=0.69\) (Fig.~\ref{fig:alignment}D). Although this restriction also increased the shuffled baseline, the observed geometry remained well above its corresponding control.

\subsection{Disentangling olfactory from other sensory and semantic structure}
\begin{figure*}[t]
    \centering
    \includegraphics[width=1\linewidth]{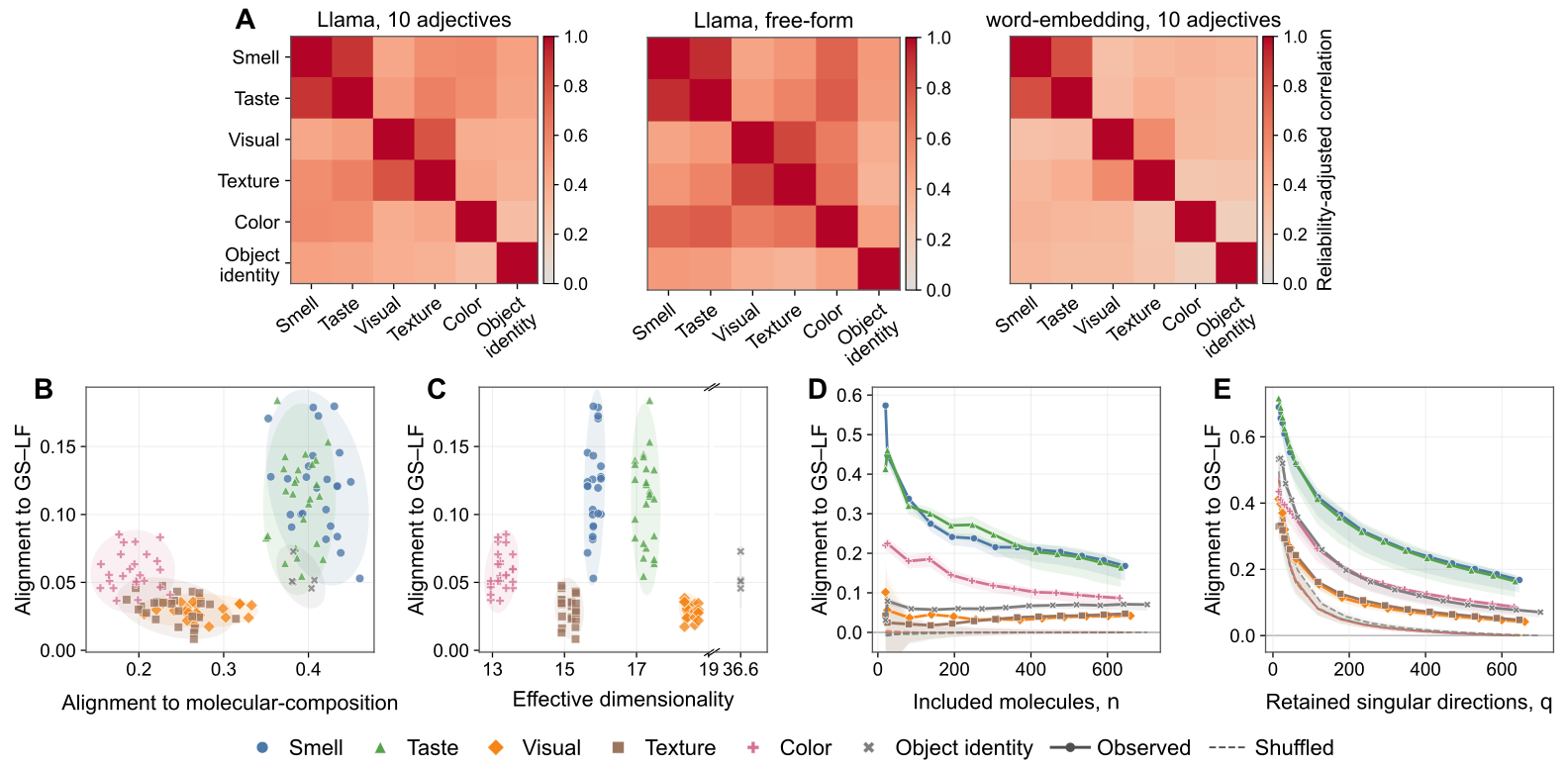}
    \caption{\textbf{Sensory organization and molecular composition alignment of language-derived percepts} 
    (A) Similarity among perceptual representations derived from different sensory prompts and mixture-name embeddings, shown for the three embedding methods. Each heatmap entry is a dimension-matched,
reliability-adjusted Pearson correlation between the corresponding
mixture RSMs.
(B) Alignment to GS--LF vs. alignment to molecular composition across
sensory and semantic targets.
(C) Alignment to GS--LF vs. the effective dimensionality of perceptual
targets, quantified using the participation ratio. Dimension matching,
reliability adjustment, and participation-ratio calculations are
described in Methods~\ref{methods:sensory_organization}.
(D,E) Alignment to GS--LF after molecular filtering and subspace restriction. These panels are similar to Fig.~\ref{fig:alignment}(C,D), but are plotted for all sensory and semantic targets (word embedding, 10 adjectives).}
    \label{fig:aggregate_senses}
\end{figure*}
We have thus far shown that smell-prompted representations contain structure that is predictable from mixture composition, and the learned molecular kernels that enabled this prediction are aligned to external odor annotations. However, olfactory perception does not occur independently of the other sensory properties of an odor source. Natural objects jointly evoke multiple of sensations such as smells, tastes and appearance, and these attributes are often correlated in both experience and language. In this section, we investigate this multisensory organization of perception.
Our database of food items is particularly suited to this analysis, as each
item can evoke sensations across multiple modalities in real-world experience.

We generated perceptual targets using prompts for smell, taste, visual appearance, texture, and color and asked whether molecular composition aligned preferentially with smell-related representations (see Appendix~\ref{app:prompt-examples} for prompting details, and Fig.~\ref{fig:top_descriptors} for top descriptors from each modality). We additionally embedded the mixture names directly, without a sensory prompt, to provide a control for generic object identity.

The perceptual target spaces exhibited substantial cross-modal organization as shown by the cross-correlation heatmaps in Fig.~\ref{fig:aggregate_senses}A. Smell and taste were particularly closely related, reaching a reliability-adjusted Pearson correlation of $\rho_{\mathrm{adj}}=0.81$
for the adjective word-embedding targets. More moderate relationships were present among other modalities including the object identity targets. For example, smell correlated with visual appearance at $\rho_{\mathrm{adj}}=0.279$
and with object identity at $\rho_{\mathrm{adj}}=0.335$
in the word-embedding space.

The above correlations were consistent across the three types of perceptual embedding.
However, the separation among modalities was clearer for the adjective word embeddings than for the Llama free-form representations. For example, the adjusted smell--visual correlation was $0.28$ for word embeddings, compared with $0.43$ for Llama activations. This difference is compatible with the broader contextual information retained by internal language-model representations in Llama, including object identity and other correlated attributes. In contrast, averaging embeddings of the requested adjectives produces a more focused representation of the prompted sensory quality since the information about the identity is already stripped.

Given the observed cross-modal correlation, we next sought to determine how our previous results on composition-alignment and learned molecular geometry changed when we use perceptual targets other than smell. For each modality, we calculated held-out alignment with mixture-composition, as well as the learned molecule-level alignment with the independent GS--LF odor annotations. We further quantified the effective dimensionality of perceptual target representations across mixtures. We found that smell and taste consistently have higher held-out alignment with mixture-compositions, as well as a higher alignment of learned embeddings to GS--LF data (Fig.~\ref{fig:aggregate_senses}B). 

Interestingly, the object-identity target, derived from mixture-name embeddings, shows a comparable  mixture-composition alignment as smell and taste. It also has a substantially higher effective dimensionality than other modalities. 
The higher dimensionality can be explained by the fact that prompting for a particular sensory modality restricts the representation to a narrower subspace that is less rich than that of the object identity. As a consequence, objects that differ in many aspects may be represented similarly when they share a modality-specific qualities, leading to a compressed variation. Object identity targets, however, lead to molecular alignment to GS--LF that is significantly lower than the smell and taste (Fig.~\ref{fig:aggregate_senses}C). This suggests the pure information content cannot solely explain external odor alignment, and rather the nature of the supervision also matters.

Finally, we extended the molecule-stability and observable-subspace analyses described in previous section to all modalities. 
Retaining the most geometrically stable molecules increased smell and taste alignment, and to a moderate degree color, and left other modalities unchanged (Fig.~\ref{fig:aggregate_senses}D). Further, while subspace restriction elevated the shuffled baseline, smell and taste alignment remained clearly above its corresponding control, while other nonchemosensory and object-identity targets staying lower and approaching the chance at small subspace dimensions (Fig.~\ref{fig:aggregate_senses}E). Thus, accounting for uneven identifiability strengthened the chemosensory ordering rather than producing a generic increase across all target types.

Overall, we see that despite cross-correlations among sensory and semantic targets, chemosensory prompting appears to emphasize aspects of the perception whose geometry has a higher alignment with that of the volatile molecular composition. 
Additionally, we found that the external odor alignment is not simply a consequence of learning any target that is predictable from composition. Rather, chemosensory supervision preferentially recovers molecule-level relationships that generalize to independently annotated odor qualities.

\subsection{Predicting mixture perception from external molecular features}

\begin{figure*}[t]
    \centering
    \includegraphics[width=0.55\linewidth]{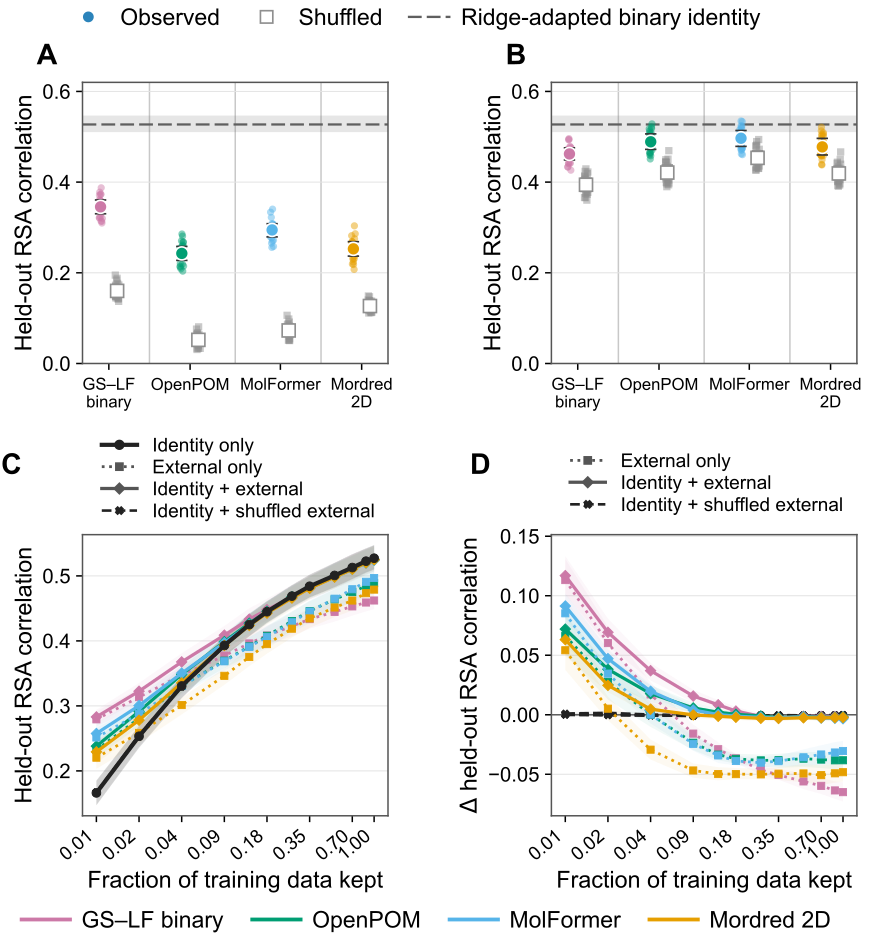}
    \caption{\textbf{Predicting mixture perception from external molecular features} 
    (A) Zero-shot alignment between mixtures created from features from GS–LF binary, OpenPOM, MolFormer, or Mordred 2D, and perceptual geometries.
    Mixtures created from all features demonstrate higher than chance alignment to perceptual geometry, but their alignment is significantly lower than the one learned from binary molecular compositions (dashed line). (B) Ridge-adapted alignment using external features only. Gray bars show molecule-row-shuffled controls. Error bars and shaded bands indicate $95\%$ crossed-bootstrap confidence intervals; dashed lines show learned \(E\) in A and ridge-adapted binary identity in B.
    (C) Held-out mixture-perception RSA for identity-only, external-only, identity-plus-external, and identity-plus-shuffled-external models. (D) The corresponding change in RSA relative to the identity-only model. Shading indicates crossed-bootstrap 95\% confidence intervals across five runs and five data splits. All data in this figure are based on the Llama 10-adjective target representation.
    }
    \label{fig:external-mixture-prediction}
\end{figure*}
Our predictions so far were agnostic to prior knowledge about molecular
identity. The molecules were treated as having arbitrary labels, and no information about their chemical structure or known odor
properties were used.
A natural question is whether such information contains any predictive power for mixture perception. To examine this, we gathered external features, including GS--LF odor features \cite{lee2023pom}, OpenPOM \cite{lee2023pom, openpom2023}, MolFormer \cite{ross2022molformer}, and Mordred descriptors \cite{moriwaki2018mordred}, for a portion of the molecules in our database (approximately \(2{,}400\) unique molecules) and used these features in our model.
We first
evaluated them zero-shot, without using any mixture-level perceptual data or
supervision.

Although their performance was substantially lower
than that of the learned molecular embedding ($r=0.53$), all
representations aligned significantly above molecule-permuted controls, with
held-out correlations of approximately $r=0.24$--$0.35$
(Fig.~\ref{fig:external-mixture-prediction}A).
Hence, external features assigned
to individual molecules contain perceptually relevant information that
transfers directly to natural mixtures.

We next considered a supervised setting, where we used external features as input instead of binary molecular identity. We trained ridge-adapted models on the the mixture representations resulting from these features (see Appendix \ref{appendix:external_features} for details). These models
reached held-out correlations of approximately $r=0.45$--$0.49$ which is close but still below the performance of ridge trained on binary molecular composition directly as well as the low-rank kernel model
(Fig.~\ref{fig:external-mixture-prediction}B).
However, note that after training the molecule-permuted shuffles also perform competitively, although still at a lower performance than the observed features. This is primarily because a sufficiently high-dimensional random feature has enough capacity to preserve the geometry of the mixture composition, hence enabling supervised prediction of perceptions (see Fig.~\ref{fig:randm_features} for a comparison to random Gaussian features of different dimensions).

The above analysis establishes that mixture perception can be learned
from these external representations, but it still does not determine whether they add
predictive information beyond that already available from molecular identity.
To test their incremental value directly, we designed a separate experiment in which models received either binary molecular-identity vectors indicating the presence or absence of each component, external molecular features, or both representations together. 
Importantly, we performed these in a budgeted training experiment to determine how the value of external molecular features changes with the amount of available perceptual supervision.
In this setup, the ridge
models were trained using perceptual targets from progressively larger subsets
of the training mixtures (Appendix \ref{appendix:external_features}). The results show that with only a small number of labeled mixtures,
combining identity with external features substantially
outperformed identity alone, whereas adding molecule-permuted features produced
no improvement (Fig.~\ref{fig:external-mixture-prediction}C,D). As it can be seen in Fig.~\ref{fig:external-mixture-prediction}D, this advantage
decreased as additional perceptual data became available and disappeared with
the complete training set. External features therefore did not provide
detectable predictive information beyond molecular identity in the full-data
regime. Instead, they supplied a useful inductive bias in the data-limited
regime. In this regime, the model was able to use pre-existing chemical or odor relationships among molecules before those relationships could
be learned directly from the mixture-perception data.

\section{Discussion}
In this work, we studied the problem of predicting perceptual similarity among complex natural odor mixtures. To do this, we used a large dataset of natural mixtures from an existing volatile-composition database and augmented it with LLM-derived surrogate perceptual representations. This combined composition--perception dataset enabled us to learn mixture-similarity metrics that captured relationships beyond simple molecular overlap.

In learning mixture similarity, we supplied the model with no molecular descriptors or prior information about chemical structure or odor quality. Molecules were represented only by their identities as distinct columns in the binary composition matrix. The held-out RSA correlation measures how well the model recovered the relative perceptual organization of mixtures that were not used for training or model selection. The above-chance performance of the identity kernel shows that direct component overlap already carries perceptually relevant information, while the improvement obtained by learning molecule-specific diagonal weights shows that components differ in how informative they are. The substantially stronger performance of the low-rank kernel indicates that generalization also benefits from learning relationships among distinct molecules (Fig.~\ref{fig2}). 
This is evidenced most directly by mixture pairs with no components in common, in which the learned low-rank kernel retained a held-out correlation of $r=0.432$. 
This suggests that the model was able to infer that different molecules can play similar roles in organizing the language-derived perceptual space.

The learned molecular geometry also aligned above chance with independent human odor annotations that were not used during training. This suggests that, for at least part of the molecular space, relationships among the learned embeddings capture perceptual relationships among the corresponding molecules. The stronger alignment obtained from chemosensory targets further suggests that this correspondence reflects odor-relevant structure rather than a generic consequence of fitting any language-derived target.

Recent studies suggest that mixture perception can often be predicted from informative representations of the constituent molecules. Perceptual profiles of controlled mixtures can be approximated by averaging the profiles of their components \citep{pellegrino2026odors}, while molecular representations combined with mixture-appropriate inductive biases can predict perceptual similarity from limited data \citep{tom2025mixtures}. Similarly, odor-informed molecular features were particularly effective for predicting human-rated mixture distances in the DREAM challenge \citep{satarifard2026semantic}. These findings motivate a further question: how much do external molecular representations contribute beyond binary composition, and how does their contribution depend on the amount of mixture-level supervision? In our experiments, external molecular representations predicted language-derived mixture similarities to a modest level even without mixture-level training (zero-shot). During supervised training, adding these features to binary composition provided the largest improvements when training data were limited (Fig.~\ref{fig:external-mixture-prediction}C,D). However, their incremental advantage diminished as more training data became available. Overall, these results support component-level representations as a useful inductive bias especially when perceptual measurements are limited, while extending this principle to the language-derived perceptual organization of complex natural odors. Finally, the predictive success of our additive representation does not imply that percept formation itself is necessarily linear. The learned component embeddings may still be flexible enough to approximate some of the perceptual effects of nonlinear interactions over the observed distribution of mixture compositions.

Recent work has begun to explore the use of language and vision-language models for odor generation \citep{wen2026aromagen} and for linking olfactory signals with visual and textual representations \citep{tsonis2026images}.
LLMs have also been evaluated for their knowledge of the perceptual qualities of individual odorant molecules \citep{makri2026benchmark}. Our results suggest an important distinction between such single-molecule queries and descriptions of natural odor sources. We found that LLM responses for individual molecules were less consistent across generations than those for natural mixtures (Fig.~\ref{fig:prompt_stability}). Natural odor sources may be represented across a richer range of contexts in language-model training data, allowing their representations to draw on distributed sensory and semantic associations rather than primarily on lookup-like associations between chemical names and odor labels. This richness may make natural mixtures particularly suitable for constructing stable surrogate perceptual geometries, although it may also introduce information that is not exclusively olfactory.

Our study has some limitations. Because concentration data were not consistently available across entries, we represented mixtures using binary component lists. This omits concentration-dependent changes in the odor qualities of individual molecules and their contributions to a mixture. The entries in the VCF data are also compiled across studies with different experimental conditions and may provide incomplete descriptions of components in the natural sources. Nevertheless, the generalization observed here suggests that large collections of natural-mixture data can support predictive modeling even when their characterization is imperfect. This motivates future efforts to collect large-scale datasets pairing controlled chemical composition with perceptual measurements of odor mixtures.

The learned mapping from molecular composition to language-derived perceptual representations could guide the design of mixtures with desired odor qualities. In principle, a desired odor description could be embedded in the perceptual space to define a target direction relative to an existing mixture. An inverse mapping could then identify direction in the composition space that correspond to that direction. This will determine component additions, removals, or substitutions to steer the mixture toward that target. However, one can imagine different qualities may vary in their molecular specificity, as some may depend on a small set of components, whereas others may be achievable through a broader range of molecules and alternative compositions. Quantifying perceptual space and its relationship to molecular composition could therefore provide a foundation for both understanding mixture perception and deliberately shaping it.

In conclusion, our study provides a scalable framework for learning
the perceptual organization of complex natural mixtures, using
language-derived odor descriptions to supervise prediction from
molecular composition.
This approach expands the scope of quantitative mixture modeling
and provides a basis for investigating how changes in composition
shape odor perception.

\section{Methods}
\label{sec:app_methods}

\subsection{Natural-mixture composition data}
The natural mixture-component matrix is constructed from VCF Online, a curated, subscription-based database of volatile compounds reported in natural and processed food products \citep{vcfonline2026}. Each entry corresponds to a food source or product entry, and contains the volatile molecular component in it. From this data, a binary matrix of component occurrences are created. After filtering of duplicates and invalid entries, we construct a matrix with 1,240 mixtures and and 8,503 molecular components. Additional target-specific
quality control was applied after language-model generation, so the exact number of mixtures varied slightly across target
constructions.

\subsection{Language-derived perceptual targets}

As mentioned in the Results (Section~\ref{ResultsA}), we generate three perceptual targets by prompting \texttt{Llama-3.1-8B-Instruct}. This includes the word-embedding and Llama internal activations (both prompted for adjectives), as well as Llama free-form text generation. A complete list of prompts and example model outputs is shown in Appendix \ref{app:prompt-examples}. We selected \texttt{Llama-3.1-8B-Instruct} because its openly available weights enabled direct access to internal activations, while its instruction tuning made it suitable for sensory-specific
prompts. Further, its relatively modest size made repeated generation and layer-wise representation analyses computationally tractable. As a default, the model was asked to return 10 adjectives, but we also swept for $\{3,5,7,10,15,20,30,50\}$ adjectives. The alignment results associated with these adjectives are shown in Fig.~\ref{fig:adjective-count-curves}.
For each perceptual target and each condition, we generated five independent responses for each mixture, which we used to asssess uncertainty.

We constructed perceptual vectors in two complementary ways. First, we
extracted the residual-stream activations after transformer block 18 for the
tokens generated in the response (we chose layer 18 as a default, however, see Fig.~\ref{fig:llama-layer-rsa} for a sweep of all layers). These 4096-dimensional activations were
averaged across generated tokens to form one vector for each mixture. The same
pooling procedure was used for adjective lists and free-form descriptions. Second, we represented the generated adjective lists using pretrained
300-dimensional GloVe semantic word embeddings
(\texttt{glove-wiki-gigaword-300}) \citep{pennington2014glove}.
The vector for a mixture was then
obtained by averaging the vectors of its parsed descriptors. In cases where a descriptor contained multi words (such as "Slightly sweet"), the
embeddings of their constituent words were averaged first, before averaging across
descriptors to form a mixture representation. These cases accounted for approximately $2.7\%$ of of all descriptor occurrences.
\vspace{-1em}
\subsection{Perceptual representational similarity}
\vspace{-0.5em}
Before computing similarities, each perceptual target feature dimension was z-scored
using the mean and standard deviation estimated from the training mixtures. The same transformation was then applied to validation and test
mixtures. Features with negligible training-set variance were assigned a
scale of one. The perceptual similarity between two mixtures was then defined as the cosine similarity of the z-scored vectors.
Finally, the alignment between two
similarity matrices was quantified by the Pearson correlation between their
vectorized upper triangles.
\vspace{-1em}
\subsection{Model training and evaluation}
\vspace{-0.5em}
The low-rank representation was learned by minimizing the objective in
Eq.~\ref{eq:optimization}. The regularization term is
$\lambda R(E) \equiv
\lambda_{\mathrm{norm}}\mathcal{R}_{\mathrm{norm}}(E)
+\lambda_2\mathcal{R}_2(E)$.
The first term controls the norms of the molecular embedding vectors,
$\mathcal{R}_{\mathrm{norm}}(E)=
\frac{1}{P}\sum_{p=1}^{P}\left(\|e_p\|_2-1\right)^2$,
and the second is a normalized squared norm penalty,
$\mathcal{R}_2(E)=\|E\|_F^2/(Pk)$.
Here, $P$ denotes the number of distinct molecules appearing across all mixtures.
We used $\lambda_{\mathrm{norm}}=10^{-2}$ and $\lambda_2=10^{-5}$
in our experiments.

Parameters were initialized from a zero-mean Gaussian distribution with
standard deviation $1/\sqrt{k}$ and optimized with Adam using a learning rate
of 0.03 and gradient-norm clipping at 10. We evaluated latent dimensions
$k\in\{10,20,50,100,200,500\},$
expanding the search to values as large as $k=1500$ when validation performance
continued to improve at the upper boundary. Models were trained for up to 50
optimization steps, and both the latent dimension and training checkpoint were
selected using validation RSA. The test set was used only for final evaluation.

For the diagonal kernel, $K=\operatorname{diag}(w)$, we learned one
positive weight per molecule using the same RSA objective, with
regularization $\lambda_{\mathrm{diag}}\mathcal{R}_{\mathrm{diag}}(\theta)$,
where $\mathcal{R}_{\mathrm{diag}}(\theta)
=\frac{1}{P}\sum_{p=1}^{P}\theta_p^2$ and
$\lambda_{\mathrm{diag}}\approx10^{-3}$.
Weights were parameterized by exponentiating parameters $\theta_p$,
initialized to zero so that training started from the identity kernel.
Parameters were clipped to $[-8,8]$ before exponentiation, and the
resulting weights were rescaled to have a mean of one across all
components.
Optimization used full-batch Adam with learning rate $0.05$ for
$100$ steps, selecting the checkpoint with the highest validation RSA
among evaluations performed every five steps, including initialization.

We partition mixtures randomly into 60\% train, 20\% validation, and 20\% test sets. We used five independent but fixed split seeds to generate 5 fits per run and condition. The partition is performed by mixture rather than by pair, so train, validation, and test similarities are computed only among mixtures within the corresponding split. This yields approximately 276k train-train pairs and 31k validation or test pairs. Consequently, every reported test relation is between two mixtures that were both held out during training and model selection, and their prediction comes from the molecules that they shared with training data. If some molecules appear only in the test set and not in the training set, they are randomly initialized and carry no predictive power.

The held-out RSA correlations for the composition-kernel comparisons,
including the zero-overlap analysis (Fig.~\ref{fig2}C), and the
supplementary sensory and object-identity performance summaries were
aggregated across repeated fits.
For each prompted target, five language-model generations and five
mixture splits produced a $5\times5$ crossed design.
The reported correlations were aggregated in Fisher-$z$ space,
$
\widehat r
=
\tanh\left[
\frac{1}{25}
\sum_{g=1}^{5}\sum_{s=1}^{5}
\operatorname{arctanh}(r_{gs})
\right],
$
where $r_{gs}$ is the held-out RSA correlation for generation $g$
and split $s$.
Confidence intervals were estimated using 5,000 crossed-bootstrap
replicates, independently resampling generation and split indices
with replacement and evaluating each replicate on their Cartesian
product. For deterministic object-identity targets, correlations
were aggregated across the five splits only, and only split indices
were resampled. Reported confidence intervals are $95\%$ percentile
intervals. Shuffled controls for the held-out RSA correlations were generated by
independently permuting the perceptual labels before training. For the
low-rank and diagonal kernels, this involved retraining the models for
each permutation. The controls shown in Fig.~\ref{fig2}C are based on
10 independent permutations for a single generation run and split
combination per target condition.
\vspace{-1em}
\subsection{Alignment of molecular geometry with external features}
\vspace{-0.5em}
We compared the learned molecular geometry with the
GoodScents--Leffingwell (GS--LF) odor annotations \citep{goodscents,leffingwell} compiled by
\citet{lee2023pom}. From these annotations, we constructed two molecular
representations. First, we represented each molecule by a 138-dimensional
binary vector indicating which odor descriptors apply to it and
constructed the primary GS--LF RSM by computing cosine similarity
directly between these vectors, without centering. Second, we averaged
pretrained 300-dimensional GloVe embeddings of the descriptors assigned to each molecule
and computed cosine similarity between the resulting vectors.
Correspondence between VCF molecules and GS--LF entries was established
using SMILES representations. Of the 4,983
unique molecular structures in GS--LF, 2,418 had a matching structure
in the VCF database. After additionally requiring each molecule to
occur in at least 10 training mixtures in every included fit, the final
comparison contained 635--707 VCF molecular entries depending on the
target condition. The two GS--LF representations were compared on the
same molecule set within each target condition.

In addition to GS--LF, we also considered OpenPOM odor embeddings
\citep{lee2023pom,openpom2023}, MolFormer molecular representations
\citep{ross2022molformer}, and Mordred 2D molecular descriptors \cite{moriwaki2018mordred}
(Fig.~\ref{figS:alignment_external}). The feature dimensions were $256$ for OpenPOM, $768$ for
MolFormer, and $1171$ for Mordred 2D.
As for GS--LF, we identified matching VCF molecules and applied the
same inclusion criteria, yielding comparable numbers of retained
molecular entries. Feature
dimensions were centered across matched molecules before cosine
similarities were calculated, except for the binary GS--LF descriptors,
which were used without centering. Mordred descriptors were additionally
standardized to unit variance.

For each fitted model $f$, we constructed a molecule-level representational
similarity matrix $
R_{pq}^{(f)}
=
\cos(e_p^{(f)},e_q^{(f)})$,
where $e_p^{(f)}$ is the learned representation of molecule $p$. We defined the consensus
molecular RSM by averaging the corresponding entries across fits:
$\overline{R}_{pq}=\frac{1}{F}\sum_{f=1}^{F}R_{pq}^{(f)}$. For each generated sensory target, the consensus was computed across five
language-model generations and five mixture splits ($F=25$). For deterministic
targets, such as object identity, it was computed across the five mixture
splits ($F=5$). For each external feature space, alignment was quantified as the Pearson
correlation between the upper triangles of the consensus learned RSM and the
corresponding external RSM. Unless otherwise specified, we use this consensus RSM as a measure of alignment to external features. Additionally, the alignments of the individual
fits are shown in Fig.~\ref{fig:aggregate_senses}(B,C).

Uncertainty in alignment was estimated by crossed bootstrap resampling of language-model generations and mixture splits. Chance alignment was estimated using 100 joint row-and-column permutations of molecule identities in each external RSM (Figs.~\ref{fig:alignment} and \ref{fig:aggregate_senses}). To test whether the training procedure itself could induce alignment, we also performed a separate control in which perceptual labels were independently permuted within the training, validation, and test partitions before model fitting. We used one independent permutation for each of the 25 generation run and splits, repeated the complete low-rank fitting procedure, and calculated the alignment of the resulting consensus molecular RSM. The resulting alignment remained near zero and was comparable to that obtained from the post-training row-and-column permutations. Specifically, for the word-embedding perceptual target evaluated against the GS--LF word-embedding features, the pre-training perceptual-label shuffle yielded $r=-0.0032$ (95\% crossed-bootstrap CI $[-0.0052,\,0.0023]$), whereas the post-training row-and-column permutation control yielded a mean of $r=-0.00008$ (central 95\% null interval $[-0.00379,\,0.00442]$), indicating that the training procedure alone did not generate the observed alignment.

The alignment between learned molecular embeddings and external features were also measured after filtering molecules based on their stability across runs as well as restricting the alignment to a subspace of molecular space. This is discussed in detail in Appendix \ref{sec:identifiability}.
\vspace{-0.5em}
\subsection{Cross-sensory organization and effective dimensionality}
\vspace{-0.5em}
\label{methods:sensory_organization}
We compared perceptual representations derived from the five sensory prompts and
mixture-name embeddings separately within each embedding family
(Fig.~\ref{fig:aggregate_senses}A). These analyses used all the mixtures, and features were mean-centered across mixtures without variance
standardization. To match dimensionality across all representations, we reduced the
4096-dimensional Llama representations to
the dimensionality of the word embeddings, by retaining the leading
300 principal components. PCA was fitted separately for each condition
and generation. We then constructed mixture RSMs using cosine
similarity in these spaces. Cross-condition similarity was then quantified by the Pearson correlation
between corresponding off-diagonal upper-triangular RSM entries,
averaged over all 25 pairs of generations from the two conditions.
To account for variability across generations, we calculated
$\rho^{\mathrm{adj}}_{ij} =
\overline{\rho}_{ij}/\sqrt{R_iR_j}$,
where $\overline{\rho}_{ij}$ is the mean cross-condition correlation
and $R_i$ is the mean pairwise correlation between RSMs from the
five runs of condition $i$, quantifying the reproducibility of
its representational geometry across generations. Deterministic mixture-name representations
were assigned $R_i=1$, and comparisons involving them averaged over the five generations of the other condition.

Effective dimensionality was measured using the participation ratio,
$D_{\mathrm{PR}} =
\left(\sum_{\ell}\lambda_{\ell}\right)^2/
\sum_{\ell}\lambda_{\ell}^{2}$,
where $\lambda_{\ell}$ are the eigenvalues of the covariance matrix
of the mean-centered representations. Participation ratios were calculated separately for
each generation and once for each deterministic representation,
including the binary composition matrix. In
Fig.~\ref{fig:aggregate_senses}C, each fitted model is paired with
the participation ratio of its corresponding target generation.
\vspace{-1em}
\subsection{Predicting mixture perception from external molecular features}
\vspace{-1em}
We constructed mixture representations from GS--LF binary descriptors,
OpenPOM embeddings, MolFormer representations, and Mordred 2D descriptors
as $Z_H=AH$, where rows of $H$ contain the molecular feature vectors and $A$
is the binary composition matrix normalized by each mixture's total
component count. For zero-shot evaluation, cosine similarities between
the mixture representations were compared with perceptual similarities,
without fitting a mapping. 
For supervised prediction, we learned a linear mapping by ridge
regression, $\widehat W_\lambda =
\underset{W}{\operatorname{arg\,min}}
\left[\|\widetilde Z_H W-\widetilde Y\|_F^2
+\lambda\|W\|_F^2\right]$,
where $\widetilde Z_H$ and $\widetilde Y$ denote the centered mixture
representations and standardized perceptual targets, respectively,
restricted to the training mixtures.
The regularization
parameter was selected by maximizing validation RSA, and performance
was evaluated using cosine similarities between predicted vectors
on held-out mixtures (Fig.~\ref{fig:external-mixture-prediction}A,B).

To assess the incremental contribution of external features, We fitted ridge models using the combined mixture kernel
$G_{\eta,H}=(1-\eta)\widetilde G_I+\eta\widetilde G_H$,
where $0\leq\eta\leq1$.
Here, $\widetilde G_I$ and $\widetilde G_H$ are the centered and normalized
mixture kernels induced by $A$ and $Z_H$, respectively. This combined kernel is equivalent to concatenating the centered and
normalized composition and external-feature representations, scaled by
$\sqrt{1-\eta}$ and $\sqrt{\eta}$, respectively. Thus, $\eta=0$
corresponds to binary composition alone, $\eta=1$ to external features
alone, and intermediate values combine both representations.
Models were fitted using progressively larger subsets of the training
mixtures, while the full validation and test sets were held fixed.
The kernel weight $\eta$ and ridge penalty $\lambda$ were jointly
selected using validation RSA
(Fig.~\ref{fig:external-mixture-prediction}C,D).
Further details of normalization, fitting, and hyperparameter selection
are provided in Appendix~\ref{appendix:external_features}.
\section*{Acknowledgment}
We thank Ben Choi for assistance with data curation and scraping, and Andrew Lee for helping develop an earlier version of the LLM prompting code. We also thank Tomer Ullman, Gautam Reddy, Jacob Zavatone-Veth and Binxu Wang for their valuable feedback on the manuscript.
F.P was supported by the Harvard Center for Brain Science (CBS)-NTT Fellowship Program on the Physics of Intelligence. This work has been made possible in part by a gift from the Chan Zuckerberg Initiative Foundation to establish the Kempner Institute for the Study of Natural and Artificial Intelligence at Harvard University.
\section*{Use of generative AI}
We used large language models to assist with manuscript writing and to develop and execute data-analysis code.
\bibliographystyle{apsrev4-2}
\bibliography{references}

\clearpage

\pagenumbering{arabic}
\renewcommand*{\thepage}{S\arabic{page}}
\renewcommand{\thefigure}{S\arabic{figure}}
\setcounter{figure}{0}

\onecolumngrid
\begin{center}
    \textbf{Appendices}
\end{center}
\appendix 

\section{Prompt templates and output examples} 
\label{app:prompt-examples}

\setlength{\tabcolsep}{3pt}
\renewcommand{\arraystretch}{1.18}

\begin{scriptsize}
\begin{longtable}{@{}ll@{}}
\caption{Example prompts and model outputs for different conditions}%
\label{tab:prompt-output-examples}\\
\toprule
Condition & \multicolumn{1}{l}{Prompt / Output} \\
\midrule
\endfirsthead

\toprule
Condition & \multicolumn{1}{l}{Prompt / Output} \\
\midrule
\endhead

\bottomrule
\endfoot

Fixed 10-adjective smell &
\ttcell{0.78\linewidth}{%
\textbf{System:} You are a connoisseur and a sommelier, an expert in describing smells and flavors in exquisite detail.\\
\textbf{User:} Use 10 adjectives to describe the smell of pistachio nut (pistacia vera) (roasted). Return exactly 10 adjectives as a single comma-separated line. Do not use numbering, bullets, line breaks, explanations, or any text other than the adjectives.\\[0.3em]
\textbf{Output:} Rich, buttery, nutty, earthy, sweetly aromatic, toasted, slightly smoky, floral, herbal, complex
} \\
\addlinespace

Fixed 10-adjective taste &
\ttcell{0.78\linewidth}{%
\textbf{System:} You are a chef, gastronome, and tasting expert, skilled at describing tastes and flavors in exquisite detail.\\
\textbf{User:} Use 10 adjectives to describe the taste of pistachio nut (pistacia vera) (roasted). Return exactly 10 adjectives as a single comma-separated line. Do not use numbering, bullets, line breaks, explanations, or any text other than the adjectives.\\[0.3em]
\textbf{Output:} Nutty, rich, buttery, slightly sweet, earthy, herbaceous, slightly bitter, crunchy, toasty, nutty
} \\
\addlinespace

Fixed 10-adjective visual &
\ttcell{0.78\linewidth}{%
\textbf{System:} You are an expert observer. Provide detailed visual descriptions.\\
\textbf{User:} Use 10 adjectives to describe what the appearance of pistachio nut (pistacia vera) (roasted) looks like. Return exactly 10 adjectives as a single comma-separated line. Do not use numbering, bullets, line breaks, explanations, or any text other than the adjectives.\\[0.3em]
\textbf{Output:} Light, brown, irregular, smooth, rounded, greenish-yellow, mottled, speckled, warty, tan
} \\
\addlinespace

Fixed 10-adjective texture &
\ttcell{0.78\linewidth}{%
\textbf{System:} You are a materials expert and haptics specialist, skilled at describing textures and tactile sensations in exquisite detail.\\
\textbf{User:} Use 10 adjectives to describe the texture or feel of pistachio nut (pistacia vera) (roasted). Return exactly 10 adjectives as a single comma-separated line. Do not use numbering, bullets, line breaks, explanations, or any text other than the adjectives.\\[0.3em]
\textbf{Output:} crunchy, nutty, brittle, rough, grainy, slightly-sweet, earthy, buttery, smooth, aromatic
} \\
\addlinespace

Fixed 10-adjective color &
\ttcell{0.78\linewidth}{%
\textbf{System:} You are a color expert and visual designer, skilled at describing colors, hues, tones, and palettes in exquisite detail.\\
\textbf{User:} Use 10 adjectives to describe the typical color or color palette of pistachio nut (pistacia vera) (roasted). Return exactly 10 adjectives as a single comma-separated line. Do not use numbering, bullets, line breaks, explanations, or any text other than the adjectives.\\[0.3em]
\textbf{Output:} muted, beige, pale, golden, creamy, soft-focus greenish-beige, muted green, olive, khaki, warm
} \\
\addlinespace

Free-form smell &
\ttcell{0.78\linewidth}{%
\textbf{System:} You are a connoisseur and a sommelier, an expert in describing smells and flavors in exquisite detail.\\
\textbf{User:} Describe the smell of pistachio nut (pistacia vera) (roasted) in natural language, using no more than 80 words. Mention the most salient odor qualities, but do not force the answer into a list. Finish the response with a complete sentence and terminal punctuation.\\[0.3em]
\textbf{Output:} Roasted pistachio nut exudes a rich, velvety fragrance with a deep, earthy undertone, reminiscent of damp soil and moss. The aroma is complemented by a subtle green note, like the freshness of springtime foliage, and a nutty, toasted quality that hints at the nut's roasted depth. Overall, the scent is complex, alluring, and utterly inviting, beckoning one to savor the delight of freshly cracked pistachios.
} \\
\addlinespace

Free-form taste &
\ttcell{0.78\linewidth}{%
\textbf{System:} You are a chef, gastronome, and tasting expert, skilled at describing tastes and flavors in exquisite detail.\\
\textbf{User:} Describe the taste of pistachio nut (pistacia vera) (roasted) in natural language, using no more than 80 words. Mention the most salient taste and flavor qualities, but do not force the answer into a list. Finish the response with a complete sentence and terminal punctuation.\\[0.3em]
\textbf{Output:} Roasted pistachio nuts exhibit a rich, buttery, and nutty flavor profile, with a satisfying crunch giving way to a soft, pale green interior. The primary taste sensation is a deep, muted sweetness, balanced by a subtle saltiness and a whisper of earthy undertones. The flavor is both smooth and complex, with a lingering finish that's both mellow and satisfying. Its taste is an unforgettable culinary delight.
} \\
\addlinespace

Free-form visual &
\ttcell{0.78\linewidth}{%
\textbf{System:} You are an expert observer. Provide detailed visual descriptions.\\
\textbf{User:} Describe the appearance of pistachio nut (pistacia vera) (roasted) in natural language, using no more than 80 words. Mention the most salient visible qualities, but do not force the answer into a list. Finish the response with a complete sentence and terminal punctuation.\\[0.3em]
\textbf{Output:} A roasted pistachio nut is a small, elongated, green or beige-colored kernel with a distinctive pointed tip and a small, lighter-colored split running along the length of its surface. The surface is relatively smooth, with a slight sheen, and a network of fine lines radiating from the split. The edges may appear slightly roughened and the overall appearance is rounded at the ends, giving it a delicate, curved silhouette. It resembles a tiny, twisted seed.
} \\
\addlinespace

Free-form texture &
\ttcell{0.78\linewidth}{%
\textbf{System:} You are a materials expert and haptics specialist, skilled at describing textures and tactile sensations in exquisite detail.\\
\textbf{User:} Describe the texture or feel of pistachio nut (pistacia vera) (roasted) in natural language, using no more than 80 words. Mention the most salient textural and tactile qualities, but do not force the answer into a list. Finish the response with a complete sentence and terminal punctuation.\\[0.3em]
\textbf{Output:} Roasted pistachios have a deeply satisfying crunch that gives away to a tender, yielding flesh within. The outer shell, split open by the first bite, reveals a subtle sheen and a smooth, even surface that's just slightly tacky to the touch, like the gentle stickiness of honey. As the teeth break down the nut, the crunch dissipates, leaving behind a soft, fibrous texture that's almost silky to the tongue. This is the taste of a perfectly roasted pistachio.
} \\
\addlinespace

Free-form color &
\ttcell{0.78\linewidth}{%
\textbf{System:} You are a color expert and visual designer, skilled at describing colors, hues, tones, and palettes in exquisite detail.\\
\textbf{User:} Describe the typical color or color palette of pistachio nut (pistacia vera) (roasted) in natural language, using no more than 80 words. Mention the most salient hues, tones, and overall palette qualities, but do not force the answer into a list. Finish the response with a complete sentence and terminal punctuation.\\[0.3em]
\textbf{Output:} Roasted pistachio nuts exhibit a warm, rich, and inviting color palette that showcases a dominant earthy green tone, subtly infused with a hint of golden brown and creamy beige undertones. The overall effect is a complex, mellow shade that evokes feelings of comfort and natural elegance, with a delicate sheen that suggests a slightly oily or buttery texture---the color of roasted pistachios is reminiscent of a tranquil spring evening after a warm sunny day.
} \\
\end{longtable}
\end{scriptsize}

\clearpage

\begin{table}
\caption{Top 20 molecules ranked by empirical geometry-profile reliability for smell word-embedding 10 adjectives.}
\label{tab:smell-word2vec10-top20-geometry-reliability}
\begin{tabular}{llr}
\toprule
Molecule & SMILES & Geometry reliability \\
\midrule
(E)-3-hexen-1-ol & CCC=CCCO & 0.6525 \\
methyl 2-methylpropanoate (=methyl isobutyrate) & COC(=O)C(C)C & 0.6448 \\
(Z)-3-hexenal & CCC=CCC=O & 0.6400 \\
trimethylamine & CN(C)C & 0.6398 \\
(E)-2-hexen-1-ol & CCCC=CCO & 0.6385 \\
cis-linalool oxide (5) & C=CC1(C)CCC(C(C)(C)O)O1 & 0.6304 \\
ethyl 3-hydroxybutanoate & CCOC(=O)CC(C)O & 0.6302 \\
trans-rose oxide & CC(C)=CC1CC(C)CCO1 & 0.6296 \\
4-hydroxybenzaldehyde (=4-formylphenol) & O=Cc1ccc(O)cc1 & 0.6288 \\
tetradecanoic acid (=myristic acid) & CCCCCCCCCCCCCC(=O)O & 0.6281 \\
3-methyl-2-buten-1-ol (=prenol) & CC(C)=CCO & 0.6276 \\
trans-linalool oxide (5) & C=CC1(C)CCC(C(C)(C)O)O1 & 0.6276 \\
citronellal (=3,7-dimethyl-6-octenal) & CC(C)=CCCC(C)CC=O & 0.6259 \\
2-methoxy-4-methylphenol (=4-methylguaiacol, creosol) & COc1cc(C)ccc1O & 0.6256 \\
ethyl 2-methylpropanoate (=ethyl isobutyrate) & CCOC(=O)C(C)C & 0.6253 \\
2,6,10,14-tetramethylpentadecane (=pristane) & CC(C)CCCC(C)CCCC(C)CCCC(C)C & 0.6244 \\
geranial ($\alpha$-citral) & CC(C)=CCCC(C)=CC=O & 0.6243 \\
heptadecane & CCCCCCCCCCCCCCCCC & 0.6232 \\
methyl hexadecanoate (=methyl palmitate) & CCCCCCCCCCCCCCCC(=O)OC & 0.6225 \\
hexyl octanoate & CCCCCCCC(=O)OCCCCCC & 0.6204 \\
\bottomrule
\end{tabular}
\end{table}

\begin{figure}[t]
    \centering
    \includegraphics[width=1\linewidth]{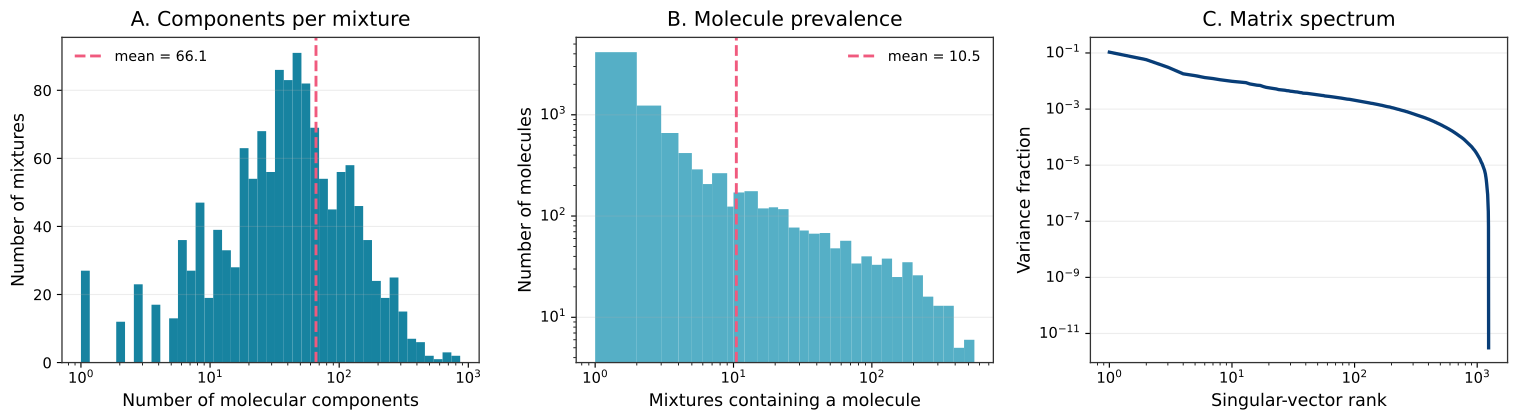}
    \caption{\textbf{VCF data statistics.} (A) Distribution of the number of components per mixtures. (B) Distribution of molecular prevalence across mixtures. Molecular prevalence is the number of mixtures a molecule appears in. (C) Singular-value spectrum of the column-centered binary mixture-by-molecule composition matrix $X$.
    }
    \label{fig:vcf_data_stat}
\end{figure}

\begin{figure}[t]
    \centering
    \includegraphics[width=1\linewidth]{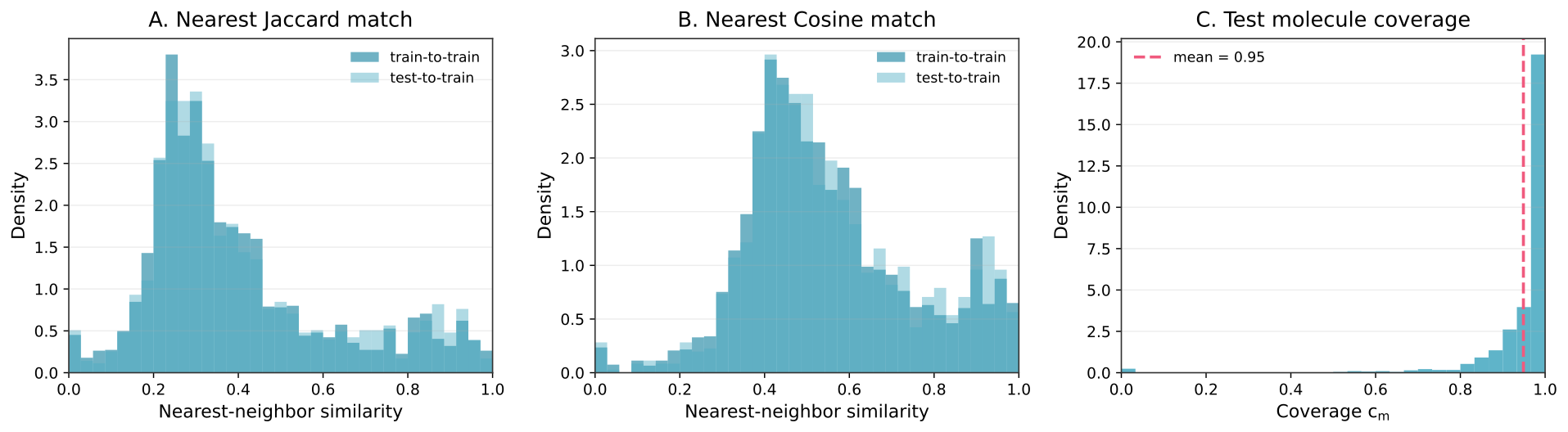}
    \caption{\textbf{Across-splits mixture coverage and similarity.} (A)-(B) Distribution of nearest neighbor similarity of train-to-train and test-to-train, as measured by Jaccard (A) and Cosine similarity (B). (C) Histogram of mixture molecular coverage between test and train. For each mixture, molecular coverage $c_m$ was calculated as $c_m = \sum_p x_{mp}\,\mathbf{1}\{p \in \mathcal{P}_{\mathrm{train}}\} \,/\, \sum_p x_{mp}.$ This measures to what extent molecules of a given mixture appeared in the training. In these plots, train/validation/test were split randomly following 60/20/20 percentiles.
    }
    \label{fig:data_splits}
\end{figure}

\begin{figure}[t]
    \centering
    \includegraphics[width=1\linewidth]{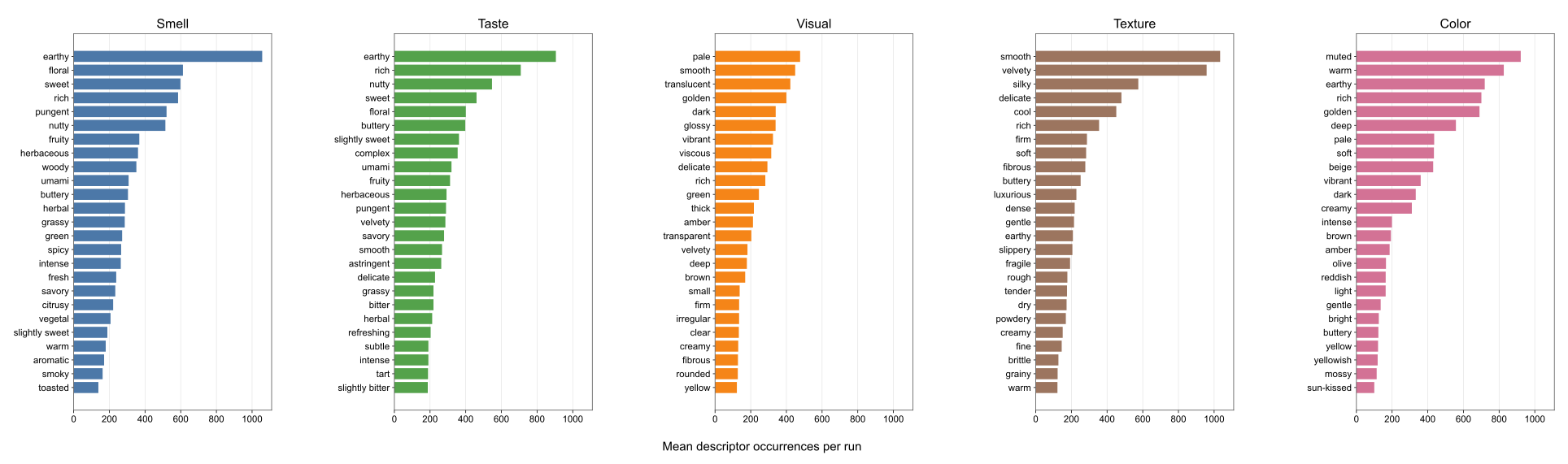}
    \caption{\textbf{Most frequent mixture descriptors across sensory prompts.} For each sensory target (10 requested adjectives), descriptor counts were pooled across five independent generation runs and divided by five.}
    \label{fig:top_descriptors}
\end{figure}

\begin{figure}[t]
    \centering
    \includegraphics[width=1\linewidth]{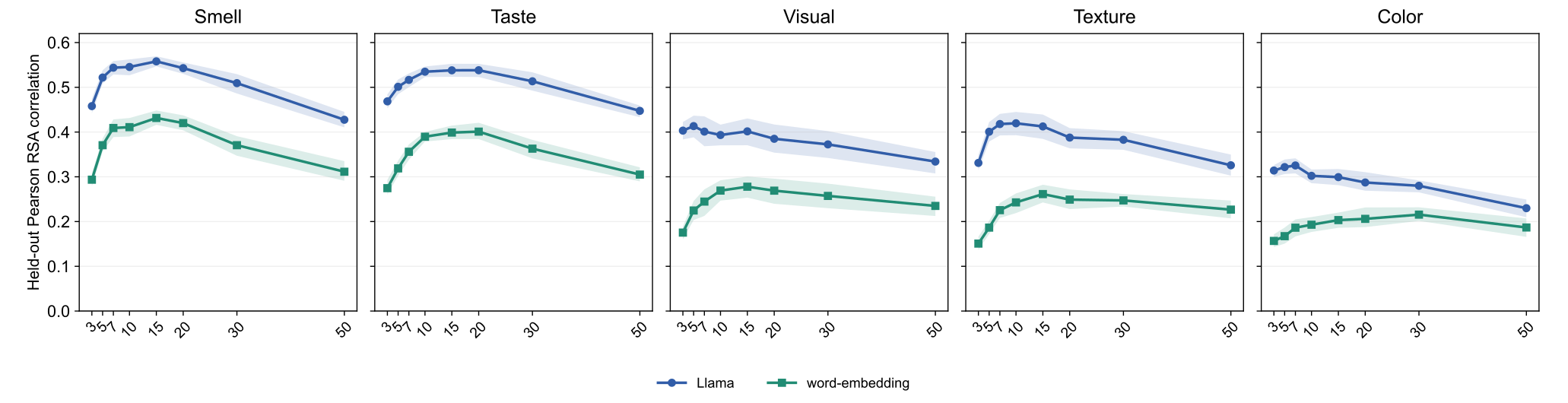}
    \caption{\textbf{Held-out RSA correlation for different numbers of prompted adjectives.}
    Curves show means over multiple split and run seeds; shaded area shows 95\% CIs.}
    \label{fig:adjective-count-curves}
\end{figure}

\begin{figure}[t]
    \centering
    \includegraphics[width=0.75\linewidth]{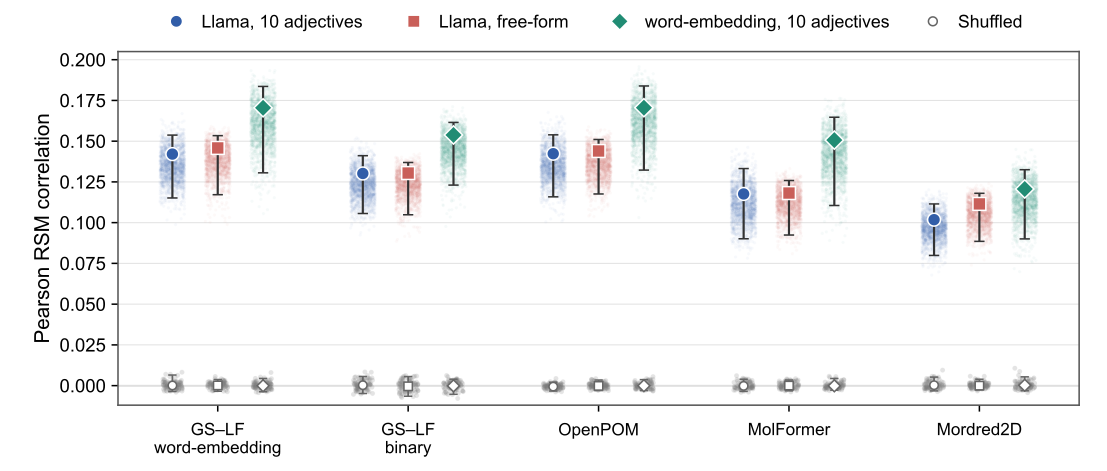}
    \caption{\textbf{ Alignment learned molecular kernels to external molecular feature spaces.}
    Bars show Pearson correlations; error bars indicate $95\%$ crossed-bootstrap CIs, and gray markers show shuffled controls. Analyses include molecules with prevalence \(\geq 10\).}
    \label{figS:alignment_external}
\end{figure}

\begin{figure}[t]
    \centering
\includegraphics[width=0.85\linewidth]{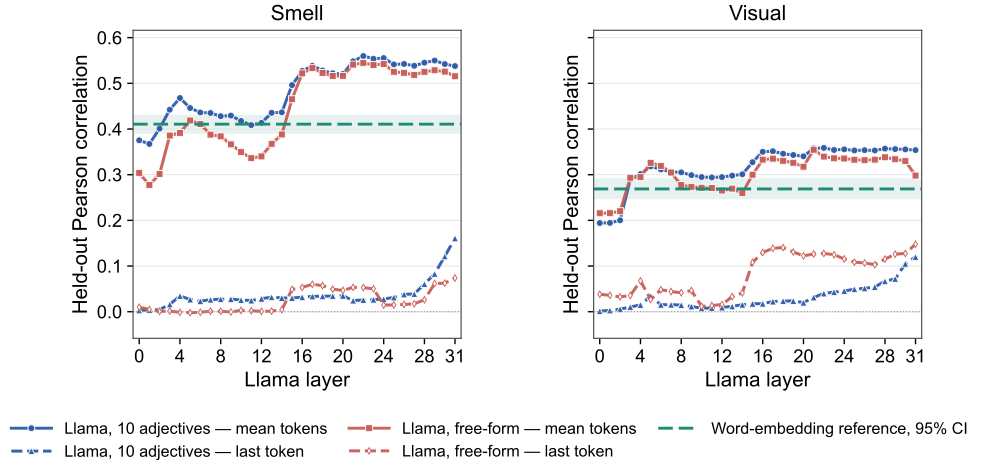}
    \caption{\textbf{Held-out composition-alignment across LLaMA layers.}
    For each given layer, the perceptual data was derived from hidden state of the network at that layer and then aligned to the molecular composition using the low-rank kernels.
Held-out Pearson RSA correlations are shown for smell (left) and visual (right) perceptual targets. Curves compare representations obtained by averaging across generated-token embeddings (solid lines) or using the final generated-token embedding (dashed lines), for 10-adjective (dark blue) and free-form (light blue) responses. The green dashed line indicates the mean performance of the 10-adjective word-embedding reference; shading denotes its 95\% crossed-bootstrap confidence interval across five generation runs and five mixture-split seeds.}
    \label{fig:llama-layer-rsa}
\end{figure}

\begin{figure}[t]
    \centering
    \includegraphics[width=0.75\linewidth]{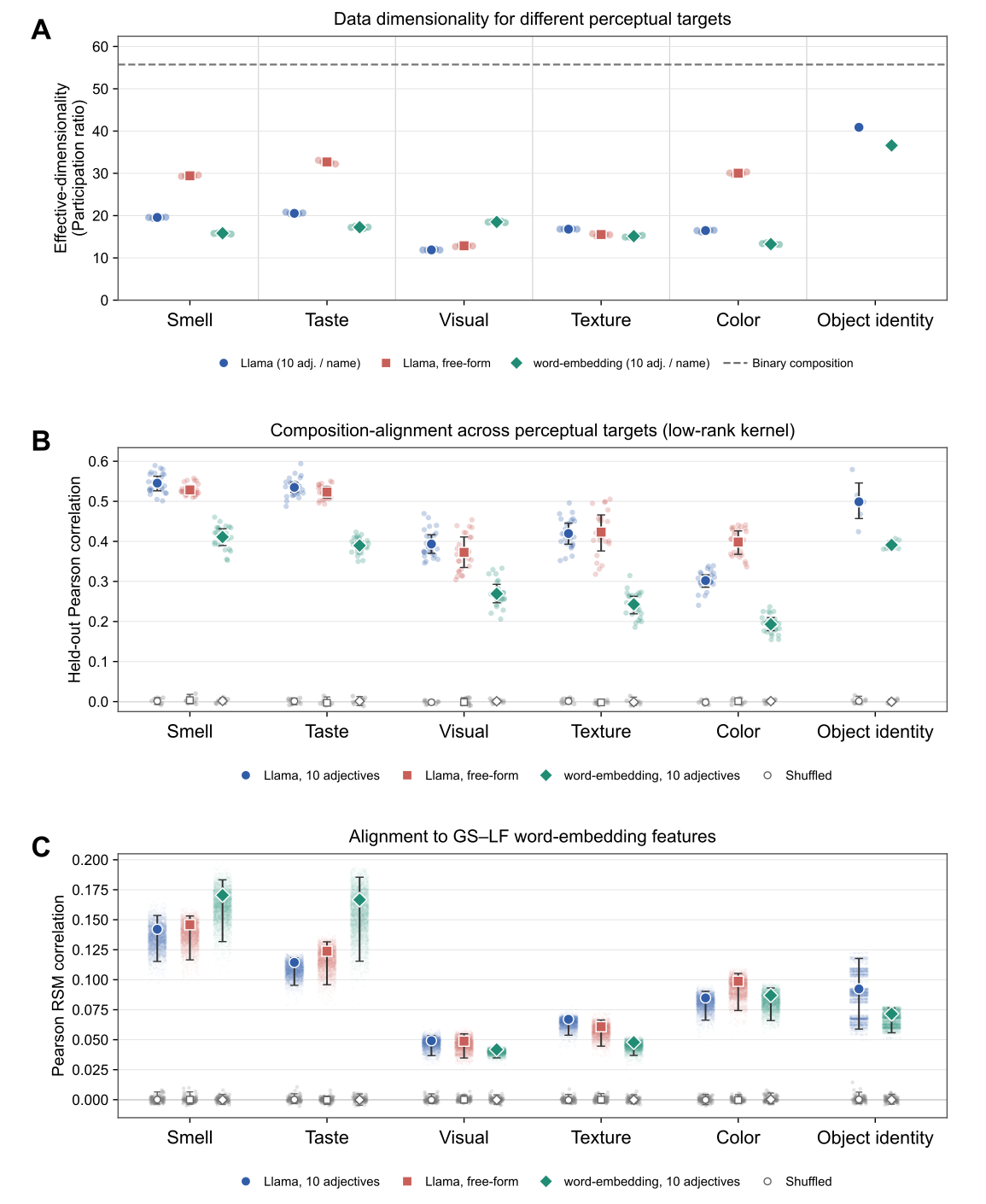}
    \caption{\textbf{Additional results on sensory organization and molecular composition alignment of language-derived percepts}
    (A) Effective dimensionality of each target representation. We quantified effective dimensionality using the participation ratio, $D_{\mathrm{PR}} = (\sum_i \lambda_i)^2 / \sum_i \lambda_i^2$, where $\lambda_i$ are the eigenvalues of the centered covariance matrix. Dashed line indicates the dimensionality of the binary mixture-composition matrix. (B) Composition alignment to each sensory target using low-rank kernels.
    (C) Alignment of learned molecular geometry to GS--LF word-embedding geometry. Bars show means and error bars indicate 95\% CIs.}
    \label{figS:aggregate_senses}
\end{figure}

\begin{figure}[h]
    \centering
    \includegraphics[width=0.35\linewidth]{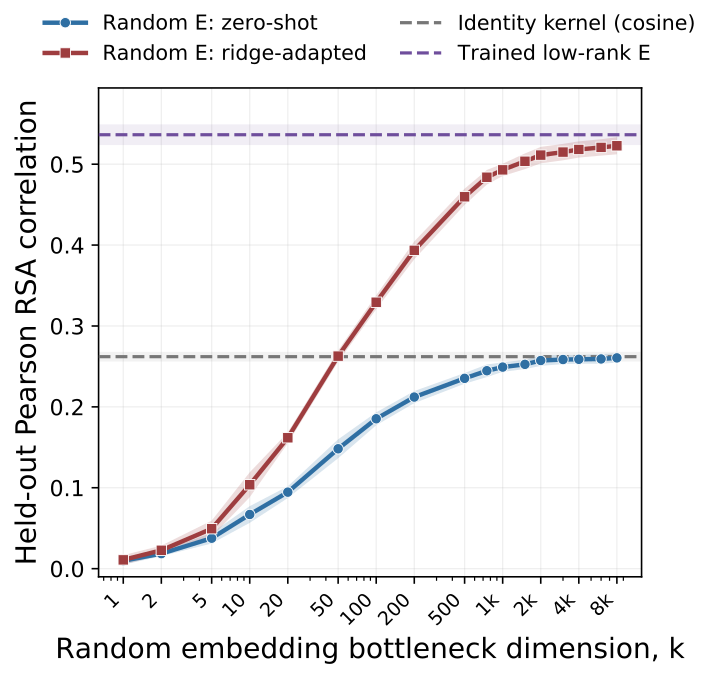}
    \caption{\textbf{Gaussian random molecular features and mixture-perception alignment} 
    Held-out RSA correlation as a function of random-feature dimension \(k\), before and after ridge adaptation. Shading denotes \(\pm 1\) SEM across five runs; dashed lines show the identity-kernel and trained low-rank-\(E\) baselines.}
    \label{fig:randm_features}
\end{figure}

\begin{figure}[h]
    \centering
    \includegraphics[width=0.7\linewidth]{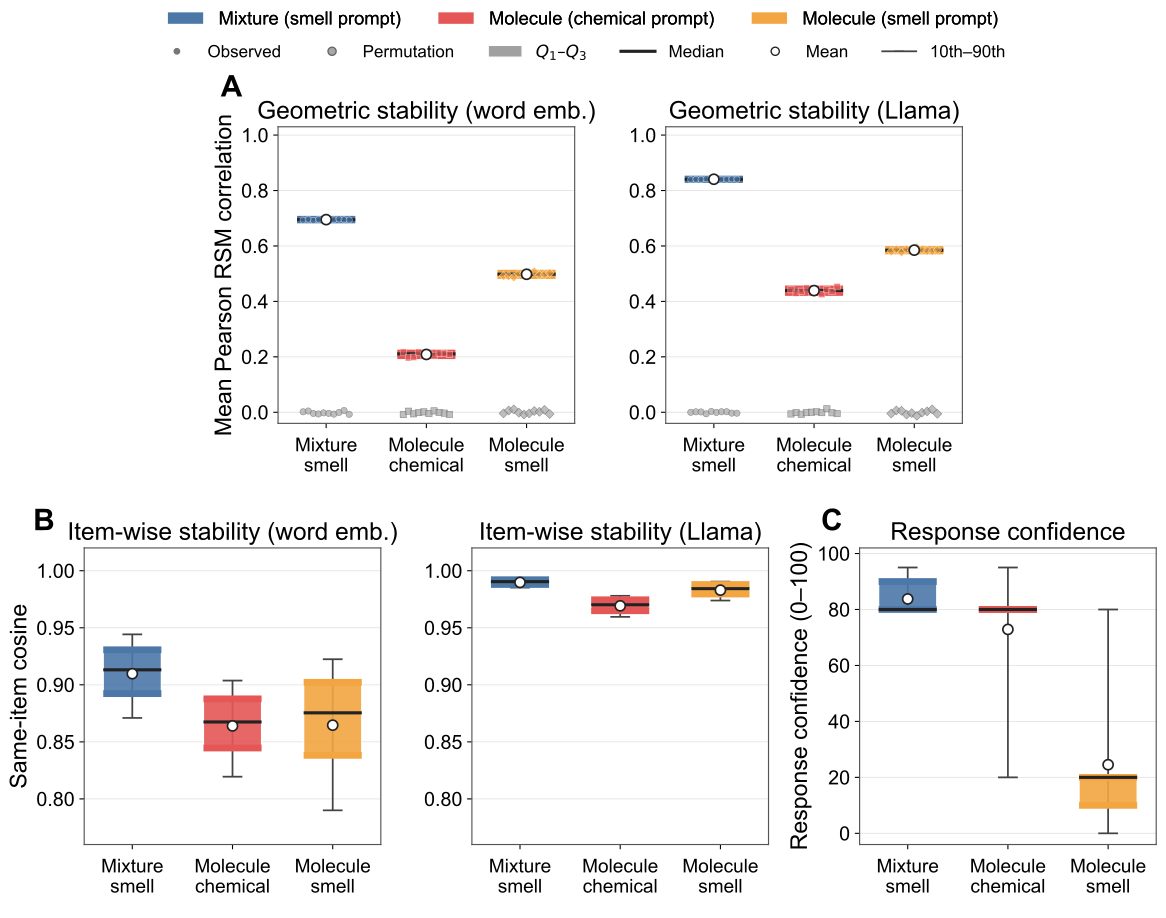}
\caption{\textbf{Prompt stability and response confidence for mixture and single-molecule prompts.}
(A) Geometric stability across prompt-generation runs, quantified as the mean Pearson correlation between representational similarity matrices constructed from word-embedding representations or Llama activations. Each colored point represents one of $B=10$ repetitions using 1,000 pair-valid items; gray points show the corresponding label-permutation controls.
(B) Item-wise stability, quantified as the mean same-item cosine similarity across valid generation-run pairs for word-embedding and Llama representations.
(C) Model-reported response confidence, averaged across the two available confidence runs. After generating the adjectives, the model received a second prompt: {\ttfamily\small\raggedright Based on your knowledge, rate your confidence that your preceding adjectives accurately describe [the target]. Use an integer from 0 to 100. Evaluate perceptual accuracy, not formatting or writing quality. Return only the integer.} 
Geometric and item-wise stability measures were calculated over 5 independent prompting runs.
Filled boxes span the interquartile range ($Q_1$--$Q_3$), black lines indicate medians, open circles indicate means, and whiskers span the 10th--90th percentiles. Colors distinguish Mixture (smell prompt), Molecule (chemical prompt), and Molecule (smell prompt).}
\label{fig:prompt_stability}
\end{figure}

\clearpage

\section{Identifiability and stability of the molecular geometry} \label{sec:identifiability}
Under the linear mixture model, 
the representations of mixtures are determined by $Z = AE$, leading to the equivalent kernel forward problem $K \rightarrow G = AKA^{\top}$. 
In this section, we discuss the stability and identifiability of estimating $K$ from observing mixture similarity matrix $G$.

\subsection{Observable subspace and recovery stability}
Let's assume $A\in\mathbb{R}^{N\times P}$, where $N$ is the number of mixtures and $P$ is the number of molecules. We assume $N<P$ and that $A$ has full
row rank $N$. Lets its compact singular-value
decomposition be
\[
A=U\Sigma V^\top,
\qquad
\Sigma=\operatorname{Diag}(\sigma_1,\ldots,\sigma_N),
\qquad
\sigma_1\geq\cdots\geq\sigma_N>0,
\]
where $V=[v_1,\ldots,v_N]\in\mathbb{R}^{P\times N}$. 
The projector onto the row space of $A$ is
\[
\Pi_A=A^\dagger A=VV^\top.
\]
First, we see that for any kernel $K$, the observable part of it is
\[ K_{obs} = \Pi_A K \Pi_A.\]
This can be seen by defining $\Pi_A' = I - \Pi_A$ and decomposing $K$ as
\begin{align*}
    K &= (\Pi_A+\Pi'_A)K(\Pi_A+\Pi'_A) \\
    &=
\underbrace{\Pi_AK\Pi_A}_{K_{\mathrm{obs}}}
+
\underbrace{
\Pi_AK\Pi'_A+\Pi'_AK\Pi_A+\Pi'_AK\Pi'_A
}_{K_{\mathrm{unobs}}}.
\end{align*}
From the above, we see that the unobservable part cannot affect $G$ since
\[ A K_{\mathrm{unobs}} A^{\top} = 0.\]
Hence, the component of $K$
supported on the null space of $A$ is fundamentally
unidentifiable from mixture observations, irrespective of the estimation procedure.

Identifiability alone does not guarantee stable recovery. This is because even within the observable subspace, some directions are much more
weakly constrained by the mixture observations than others. Consider a molecular-kernel component along $v_iv_j^\top$. A perturbation of the molecular kernel along this component changes the
the mixture level kernel according to
\[
A(v_iv_j^\top)A^\top
=
\sigma_i\sigma_j u_i u_j^\top.
\]
Components associated with small singular values therefore have only a
weak effect on the observed mixture geometry and are more sensitive to
errors in the observed mixture kernel, irrespective of the estimation procedure. We can restrict the analysis to the $q$ best-conditioned molecular
directions. let
\[
V_q=[v_1,\ldots,v_q],
\qquad
\Pi_q=V_qV_q^\top,
\qquad
K^{(q)}=\Pi_qK\Pi_q,
\qquad q\leq N.
\]
Here, $K^{(q)}$ is the projection of the kernel to the $q$ most strongly observable subspace.
For any perturbation $\Delta K$ to the kernel, let
$\Delta K^{(q)}=\Pi_q\Delta K\Pi_q$.
The corresponding perturbation of the mixture kernel is
\[
\Delta G^{(q)}
=
A\Delta K^{(q)}A^\top.
\]
We have:
\[
\|\Delta K^{(q)}\|_F
\leq
\frac{1}{\sigma_q^2}
\|\Delta G^{(q)}\|_F.
\]
Since within the $q$-subspace the mapping is invertible, the above shows that $\sigma_q^{-2}$ bounds the amplification of mixture-level errors
when recovering the molecular kernel within the retained subspace. Hence, for a given level of error on the mixture side, higher $\sigma_q$ indicates a lower recovery error in subspace $q$. Decreasing $q$ removes progressively weaker directions and improves the
conditioning of the restricted problem.
In Fig.~\ref{fig:alignment}(D), we projected the estimated and reference kernels onto this same subspace and then evaluated their alignment.

Finally, note that we used the Gram matrix formalism above as a surrogate model to study the identifiability and conditioning. In practice, the RSA objective in the main text adds extra levels of non-identifiability, and hence the subspace analysis, though useful, is not exact for the original objective in Eq.~\ref{eq:optimization}. 

\subsection{Empirical molecular geometric stability}
In addition to the above subspace filtering, we also performed filtering based on a per-molecule empirical stability measure. Let $r \in [N_r]$ index generation runs, $s \in [N_s]$ index splits, $i$ index molecules, and $e_i$ be the row $i$'th row of $E$ or equivalently the embedding vector for molecule $i$. We can define the geometry profile of molecule $i$ in a given fit $(r,s)$ by its relationship to all other components as
\[ p^{(r,s)}_i = [\cos(e^{(r,s)}_i,e^{(r,s)}_j)]_{j \neq i}.\]
The empirical reliability of a molecule then can be defined by how stable is this profile across runs. This can be measured by an average of the cross-runs Pearson correlation of the profile
\begin{equation} \label{eq:reliability}
\text{Rel}(i) = \frac{1}{N_sN_r(N_sN_r-1)}\sum_{\substack{r\in [N_r] \\s \in [N_s]}} \sum_{\substack{r'\in [N_r] \\s' \in [N_s]\\ (r',s')\neq (r,s)}} \operatorname{corr}(p_i^{(r,s)}, p_i^{(r',s')}).
\end{equation}
This defines an empirical reliability metric for each molecule. For Fig.~\ref{fig:alignment}(C), molecules are ranked by
$\operatorname{Rel}(i)$, and the kernels are restricted to the largest $n$ before calculating the alignment

The two analyses above are complementary: geometric reliability filters
individual molecules according to reproducibility across repeated
estimates, whereas observable-subspace restriction filters molecular
directions according to the conditioning of the mixture design.

\section{Kernel learning and ridge regression} \label{sub:ridgeregression}
Our method for learning low-dimensional embedding is closely related to ridge regression under linear mixture model and the kernel interpretation of that. Here, for completeness, we discuss ridge regression and its application to our problem. We later apply this to external features in Appendix  \ref{appendix:external_features}.

Following the main text, let $X \in \{0,1\}^{N \times P}$ be the binary mixture composition matrix ($P$: number of all the molecules, and $N$: number of all the mixtures). With components embedding matrix
$E \in \mathbb{R}^{P \times k}$, the matrix of mixture representations becomes
\[ Z = AE \in \mathbb{R}^{N \times k},\]
where
\[A = D^{-1}X, \qquad \text{and}\qquad D = \text{Diag}(\|x_1\|_1,\cdots,\|x_N\|_1).\]
In the above, $A$ is a normalized mixture-component matrix, and $D$ contains the number of components per mixtures. The mixture Gram matrix becomes
\begin{equation}
    G = ZZ^{\top} = AKA^{\top}, \label{eq:GZZ}
\end{equation} 
where $K = EE^{\top}$ is the embedding kernel matrix.
So the forward mapping from the components to mixtures spaces can be framed in terms of their respective kernels as: \[\text{forward map:}\quad K \rightarrow G = AKA^{\top}.\]

Let $Y \in \mathbb R^{N \times d}$ contain the perceptual target vectors for the mixtures. The alignment between the mixture similarity with the perceptual similarity can be stated as an equivalence between the respective Gram matrices, yielding $$G \equiv YY^{\top}$$ (here, $Y$ could be the original or a preprocessed version of the vectors). Note that, in the main text, this equivalence was formalized by the correlation between the matrices as was maximized in Eq.~\ref{eq:optimization}. 
If we take this to be an exact equality (or a least square problem if $A$ is tall or rank-deficient), we have:
\begin{equation}
    AKA^{\top} = YY^{\top}.
\end{equation}
The minimum Frobenius-norm solution can be obtained using the pseudoinverse,
\[ \hat{K}_0 = A^\dagger YY^{\top}(A^\dagger)^\top.\]
The above kernel admits a molecular embedding $\hat{E}_{0} = A^{\dagger}Y$ that satisfies $\hat{K}_{0} = \hat{E}_{0}\hat{E}_{0}^{\top}$. Note also that in this case $E_0$ is also the minimum norm solution to $AE=Y$ equation.

A numerical problem with calculating the pseudoinverse is that small nonzero singular values $\sigma_i$ of $A$ can lead to unstable results. If $A = U\operatorname{Diag}(\sigma_i) V^{\top}$ is the singular value decomposition, then $A^{\dagger} = V\operatorname{Diag}(\frac{1}{\sigma_i}) U^{\top} = A^{\top}(AA^{\top})^{-1}$. A regularized pseudoinverse can be defined by replacing the $1/\sigma_i$ terms with $\sigma_i/(\sigma_i^2 + \lambda)$, for $\lambda>0$. This gives
\[ L_{\lambda} = V\operatorname{Diag}(\frac{\sigma_i}{\sigma_i^2 + \lambda})U^{\top} =  A^{\top}( AA^{\top} + \lambda I)^{-1},\]
and subsequently the estimated kernel becomes
\[ \hat{K}_{\lambda} = L_{\lambda}YY^{\top} L_{\lambda}^{\top}.\]
The regularized kernel admits an embedding $\hat{E}_{\lambda}$ that satisfies $\hat{K}_{\lambda} = \hat{E}_{\lambda}\hat{E}_{\lambda}^{\top}$:
\[ \hat{E}_{\lambda} = L_{\lambda}Y = A^{\top}( AA^{\top} + \lambda I)^{-1}Y.\]
This factor is the closed-form solution of the embedding-level ridge problem
$$
\hat{E}_\lambda
=\arg\min_E
\left\|AE-Y\right\|_F^2
+\lambda\lVert E\rVert_F^2.
$$

Hence, we see that the problem of equating the perceptual and the mixture component kernels is closely related to the problem of direct percept coordinate predictions. However, note that the specific recovery of kernels or the associated embeddings depends on the particular estimation or optimization procedure. For example, conditions such as low-rankness is not directly included in the above ridge problem and the embedding adopts the dimension of the perceptual vector.

\section{Using external molecular features to predict mixture perception} \label{appendix:external_features}
In the main text, we learned the molecular embedding fully from the mixture perceptual similarity data. Here, we a consider case where we have a fixed molecular feature matrix $H \in \mathbb R^{P \times d_h}$ with each row denoting the feature of a given molecule. We first describe zero-shot and ridge-adapted prediction from these features and then ask whether they provide predictive information beyond molecular identity.
\subsection{Zero-shot and ridge-adapted prediction from fixed molecular features}
 Let $A$ denote the normalized mixture-composition matrix. The representations of all mixtures under linear model becomes:
\[ Z_H = AH \in \mathbb R ^{N \times d_h}.\]
The corresponding mixture level Gram matrix is
\[
G_H
=
Z_HZ_H^\top
=
AHH^\top A^\top
=
AK_HA^\top.
\]
where 
\[ K_H = H H^{\top} \succeq 0\]
is the molecular kernel matrix induced by the external features. Note that $(K_H)_{ij} = h_i^{\top}h_j$, and similar to before cosine similarity of two mixtures $m$ and $n$ can be written as
\[ \cos(z_{H_m},z_{H_n}) = 
\frac{x_m^\top K_Hx_n}
{\sqrt{
\left(x_m^\top K_Hx_m\right)
\left(x_n^\top K_Hx_n\right)
}},\]
where
\[
x_m^\top K_Hx_n
=
\sum_{i=1}^{P}\sum_{j=1}^{P}
x_{mi}x_{nj}\,h_i^\top h_j.
\]
The similarity metric resulting from the above is the \emph{zero-shot} similarity derived from external features.

More generally, one can transform the external features using a linear adapter $W$ to get:
\[ E = HW, \quad W \in \mathbb R^{d_h \times d}.\]
The mixture representation matrix becomes $Z'_{H,W} = AE$, and the corresponding mixture Gram matrix
\[ G'_{H,W} =  Z_HWW^{\top}Z_H^{\top}.\]
One can find $W$ by optimizing for similarity matrix between $G'_{H,W}$ and the perceptual counterpart. In the case of ridge regression,
\[
\widehat W_\lambda
=
\arg\min_W
\left\{
\|Z_H W-Y\|_F^2
+
\lambda\|W\|_F^2
\right\},
\qquad
\lambda>0.
\]
Its closed-form solution is
\[
\widehat W_\lambda
=
(Z_H^\top Z_H+\lambda I_{d_h})^{-1}Z_H^\top Y.
\]

We performed two versions of external features predictions: zero-shot and ridge-adapted which are both shown in Fig.~\ref{fig:external-mixture-prediction}(A,B). In both cases, the mixture-level external features were centered, and perceptual targets were standardized using means and standard deviations estimated from the training mixtures. For the ridge adaptation result, the regularization parameter was selected by maximizing Pearson RSA on the validation mixtures and was then frozen for evaluation on the held-out test mixtures.

\subsection{Combining molecular identity with external molecular features across training budgets}

The above section performs ridge adaptation on the external features $H$.
To determine whether an external molecular representation provided predictive information beyond molecular identity, we performed additional experiment by fitting ridge models that combined the mixture-composition representation with one external molecular-feature representation at different training budgets. Let
\[
K_I=I_P
\qquad\text{and}\qquad
K_H=HH^\top
\]
denote the molecular-level identity and external-feature kernels, respectively. The corresponding mixture representations are
\[
Z_I=A,
\qquad
Z_H=AH,
\]
where any normalization used when aggregating molecules into mixtures is absorbed into \(A\). Their induced mixture-level Gram matrices are
\[
G_I=Z_IZ_I^\top=AA^\top, \qquad 
G_H=Z_HZ_H^\top=AHH^\top A^\top.
\]
We modeled the combined representation using the mixture-level kernel
\[
G_{\eta,H}
=
(1-\eta)\widetilde G_I
+
\eta\widetilde G_H,
\qquad
0\leq\eta\leq1.
\]
Here, the tilde denotes double-centering and normalization of each kernel. Specifically, for $N$ available labeled training mixtures \(\widetilde G = (1/s)CGC\), where \(C=I_N-N^{-1}\mathbf 1\mathbf 1^\top\), with the proportionality constant $s$ chosen so that the centered kernel had unit mean diagonal (the normalization constant was estimated from the complete training pool and held fixed across training budgets). The above kernel combination can be thought of as ridge regression on a weighted concatenation of the mixture-composition and external-feature:
\[
Z_{\eta,H}
=
\left[
\sqrt{1-\eta}\,\widetilde Z_I
\;\middle|\;
\sqrt{\eta}\,\widetilde Z_H
\right],
\]
where $\widetilde Z_I$ and $\widetilde Z_H$ are corresponding centered and normalized mixture representations.
To see this, it is easy to check that $
Z_{\eta,H}Z_{\eta,H}^\top
=
G_{\eta,H}$.
We normalized the two feature blocks because they had different scales, ensuring that \(\eta\) controlled their relative contributions rather than reflecting differences in their original units. The endpoints \(\eta=0\) and \(\eta=1\) correspond to molecular identity alone and external features alone, respectively.

From the above, the ridge map was estimated from the available training mixtures as
\[
\widehat W_{\eta,\lambda}
=
\left(
Z_{\eta,H}^{\top}Z_{\eta,H}
+
\lambda I
\right)^{-1}
Z_{\eta,H}^{\top}\widetilde Y.
\]
The kernel weight $\eta$ and ridge penalty $\lambda$ were selected jointly by maximizing Pearson RSA on the validation mixtures which was held fixed across training budgets.
We evaluated
$
\eta\in\{0,0.1,\ldots,1\}
$
and
$
\lambda\in\{10^{-6},10^{-5},\ldots,10^{10}\}.
$
The selected \((\widehat\eta,\widehat\lambda)\) was then frozen and evaluated once on the held-out test mixtures. This procedure was performed for each external feature family separately. Finally, for shuffled controls, the molecular rows of \(H\) were independently permuted in each shuffle replicate:
\[
Z_H^{(b)}=AP_bH,
\qquad b=1,\ldots,B,
\]
where \(P_b\) denotes the permutation matrix for the \(b\)-th shuffle. The complete kernel construction, normalization, ridge fitting, and validation-selection procedure was repeated for every permutation.
\end{document}